\documentclass[aos,preprint]{imsart}

\RequirePackage[OT1]{fontenc}
\RequirePackage{amsthm,amsmath,bm,amssymb}
\RequirePackage{natbib}
\RequirePackage[colorlinks,citecolor=blue,urlcolor=blue]{hyperref}

\RequirePackage{multirow,booktabs}
\RequirePackage{enumitem}
\RequirePackage[usenames,dvipsnames]{xcolor}
\RequirePackage{graphicx,subfigure}
\usepackage{ulem}
\usepackage{xr}
\usepackage{cleveref}
\arxiv{arXiv:0000.0000}

\startlocaldefs
\numberwithin{equation}{section}
\theoremstyle{plain}
\newtheorem{thm}{Theorem}[section]

\newtheorem{lam}{Lemma}[section]

\newcommand{\boldG}{\mbox{\boldmath$\Gamma$}}

\endlocaldefs

\usepackage{bm}

\begin{document}
\begin{frontmatter}
\title{Model selection for high-dimensional linear regression with dependent observations}

\begin{aug}
\author{\fnms{Ching-Kang} \snm{Ing}\thanksref{t1}\ead[label=e1]{cking@stat.nthu.edu.tw}}

\thankstext{t1}{Research was supported in part by the Science Vanguard Research Program of the Ministry of Science and Technology, Taiwan}

\runauthor{C.-K. Ing}

\affiliation{National Tsing Hua
University}

\end{aug}

\begin{abstract}
We investigate the prediction capability of the orthogonal greedy
algorithm (OGA) in high-dimensional regression models with dependent observations.
The rates of convergence of the prediction error of
OGA are obtained under a variety of sparsity conditions.
To prevent OGA from overfitting, we introduce a high-dimensional Akaike's information criterion
(HDAIC) to determine the number of OGA iterations. A key contribution of this work is to show that OGA, used in
conjunction with HDAIC, can achieve the optimal convergence rate without knowledge
of how sparse the underlying high-dimensional model is.
\end{abstract}



\end{frontmatter}

\section{Introduction}

\indent
Model selection for high-dimensional regression models
has been one of the most vibrant topics in statistics over the past decade.
It also has broad applications in a variety of
important fields such as bioinformatics,
quantitative finance, image processing, and advanced manufacturing; see Negahban et al. (2012)
and Ing et al. (2017) for further discussion.
A typical high-dimensional regression model
takes the following form:
\begin{eqnarray}
\label{aug1}
y_t=\sum_{j=1}^{p} \beta_{j}x_{tj}+\varepsilon_t, \,\,t=1,\ldots,n,
\end{eqnarray}
where $n$ is the sample size, $x_{t1}, \cdots, x_{tp}$ are predictor variables,
$\varepsilon_{t}$ are mean-zero random disturbance terms, and
$p=p_n$ is allowed to be much larger than $n$.
There are
computational and statistical difficulties in estimating the
regression function by standard regression methods owing to $p \gg n$.
However, by assuming sparsity conditions on $\beta_j$,
eigenvalue conditions on the covariance (correlation) matrix of the predictor variables,
and distributional conditions on $\varepsilon_t$ or $x_{tj}$,
it has been shown that
consistent estimation of the regression function
or optimal prediction is still possible either through penalized least squares methods
(see Zhao and Yu (2006), Candes and Tao (2007), Bickel et al. (2009), and Zhang (2010))
or through
greedy forward selection algorithms (see B\"{u}hlmann (2006),
Chen and Chen (2008), Wang (2009), Fan and Lv (2008), and Ing and Lai (2011)).

The vast majority of studies on model \eqref{aug1}, however, have focused on
situations where $\mathbf{x}_{t}=(x_{t1}, \ldots, x_{tp})^{\top}$
are nonrandom and $\varepsilon_t$ are independently and identically distributed (i.i.d.)
or $(\mathbf{x}_{t}, \varepsilon_t)$ are i.i.d., which regrettably preclude most serially correlated data.
In fact, \eqref{aug1} can encompass a broad array of
time series models if these restrictions are relaxed.
For example, it
becomes the well-known autoregressive (AR)
model when $x_{tj}=y_{t-j}$.
Since the predictor variables
in AR models have a natural ordering, a commonly used sparsity condition
is
\begin{eqnarray}
\label{intro1}
C_1j^{-\gamma} \leq |\beta_{j}| \leq C_2j^{-\gamma},\,\, 0<C_1\leq C_2<\infty, \gamma>1,
\end{eqnarray}
in which $|\beta_{j}|$ decay polynomially, or
\begin{align}
\label{intro2}
C_3\exp(-\beta j) \leq |\beta_{j}| \leq C_4 \exp(-\beta j), \,\,0<C_3\leq C_4<\infty, \beta>0,
\end{align}
in which $|\beta_{j}|$ decay exponentially (see Shibata (1980) and Ing (2007)).
Moreover, the model selection problem in the AR case is simplified to an order selection one,
which has been well explored in the literature (see Shibata (1980)).
When $x_{tj}, j=1, \ldots, p$, do not have a natural ordering,
e.g., the autoregressive exogenous (ARX) model,
\eqref{intro1} and \eqref{intro2} can be generalized as
\begin{align}
\label{ts5}
L j^{-\gamma} \leq |\beta^{*}_{(j)}| \leq Uj^{-\gamma},
\end{align}
and
\begin{align}
\label{intro3}
L_1 \exp(-\beta j) \leq |\beta^{*}_{(j)}| \leq U_1 \exp(-\beta j),
\end{align}
respectively, where $0<L\leq U<\infty$,
$0<L_1\leq U_1<\infty$, and $|\beta^{*}_{(1)}| \geq |\beta^{*}_{(2)}|\geq \cdots \geq
|\beta^{*}_{(p)}|$ is a rearrangement of $\{|\beta^{*}_{j}|\}$
in decreasing order
with $\beta^{*}_{j}=\sigma_j\beta_j$
and $\sigma_j^2=\mathrm{E}(x_{tj}^2)$.
However,
unlike the order selection problem,
the model selection problem in \eqref{aug1}
with dependent observations and with coefficients satisfying \eqref{ts5} or \eqref{intro3}
seems to be seldom investigated.
The problem becomes more challenging when
$\beta_j$ may obey either one of
\eqref{ts5}, \eqref{intro3}, or $k_0 \ll n$,
but it is unclear which of the three is true.
Here, $k_0$ denotes the number of nonzero coefficients in model \eqref{aug1},
and $k_0 \ll n$ is referred to as the strong sparsity condition.

In this paper, we assume that
the $(\mathbf{x}_t, \varepsilon_t)$ in model \eqref{aug1}
is a time series obeying
concentration inequalities \eqref{ts1.1}
and \eqref{ts2.1}. We also assume that
the $\beta_j$ in model \eqref{aug1}
follow one of the following sparsity conditions:
(i) (A3), (ii) (A4), or (iii) $k_0 \ll n$,
where (A3) and (A4) are defined in Section \ref{sec2.1}.
Note that (A3) includes \eqref{ts5} and
\begin{eqnarray}
\label{ts4}
\sum_{j =1}^{p}|\beta_{j}^{*}|^{1/\gamma}< M_4, \,\,\mbox{for some}\,\,\gamma\geq1,\,\, 0<M_4<\infty,
\end{eqnarray}
as special cases, whereas (A4) contains \eqref{intro3}.
We use the orthogonal greedy algorithm (OGA) (Temlyakov, 2000)
to sequentially include candidate variables
and introduce a high-dimensional Akaike's information criterion
(HDAIC) 
to determine the number of OGA iterations.
This model selection procedure is denoted by OGA+HDAIC.
A key contribution of this paper is to show that OGA+HDAIC achieves the optimal convergence rate without
knowing which sparsity condition among (i), (ii), and (iii) would follow, thereby
alleviating the dilemma mentioned in the previous paragraph.

Following this introduction, the rest of the paper is organized as follows.
In Section \ref{sec2.1}, we introduce OGA and
the assumptions required for our asymptotic analysis of the algorithm.
Section \ref{sec2.3} derives an error bound for OGA, which is the
sum of an approximation error and a term accounting for the sampling
variability. Since the approximation error decreases as the number
$m$ of iterations increases and the sampling variability increases
with $m$, the optimal $m$ can be determined by equating the two
terms in the error bound for OGA. This approach, however, is
infeasible because not only does the solution involve the unknown
parameters in (A3) or (A4), but it is unknown which kind of sparsity
among (i), (ii), and (iii) holds true. To overcome this difficulty,
Theorem \ref{thm3.1} in Section \ref{sec3.1} proposes using HDIC to
determine the number of iterations, and shows that OGA+HDAIC is rate
optimal regardless of which sparsity condition is true. In Section
\ref{sec3.11}, we offer a comprehensive comparison of our results
with those in Negahban et al. (2012) and Ing and Lai (2011), in
which the statistical properties of Lasso (Tibshirani, 1996) and
OGA, respectively, are explored under model \eqref{aug1} with independent
observations. In this connection, Section \ref{sec3.11} also
discusses the papers by Basu and Michailidis (2015) and Wu and Wu
(2016), which investigate the performance of Lasso under sparse high-dimensional
time series models. The proof of Theorem \ref{thm3.1} is given in Section
\ref{sec3.2}. We conclude in Section \ref{sec4}. An appendix
consisting of some technical results is given at the end of the
paper. A simulation study to illustrate the performance of
OGA+HDAIC, along with further technical details, is deferred to the
supplementary material.


\section{Asymptotic Theory of OGA in Weakly Sparse Models}
\label{sec2}
\indent
This section aims at establishing the convergence rate of OGA under
sparse high-dimensional regression models with dependent observations.
The definition of OGA and the assumptions required for our analysis of OGA
are given in Section \ref{sec2.1}.
The main result of this section is stated and proved in Section \ref{sec2.3}.

\subsection{Models and Assumptions}
\label{sec2.1}
We assume that
$\{(\mathbf{x}_{t}, \varepsilon_t)\}$
in model \eqref{aug1} is a zero-mean stationary time series
satisfying $\mathrm{E}(\mathbf{x}_{t}\varepsilon_t)=\mathbf{0}$.
The OGA is a recursive procedure that selects variables from
the set of predictor variables in \eqref{aug1} one at a time.
Define
$\mathbf{X}_{i}=(x_{1i},\ldots, x_{ni})^{\top}$,
$\mathbf{Z}_{i}=(z_{1i},\ldots, z_{ni})^{\top}=\mathbf{X}_{i}/\sigma_i$,
and
$\mathbf{Y}=(y_{1}, \ldots, y_{n})^{\top}$.
The algorithm is initialized by setting 
$\hat{J}_{0}= \emptyset$, where $\hat{J}_m$ denotes the index set of
the variables chosen by OGA at the $m$-th iteration.
For $m \geq 1$, 
$\hat{J}_{m}$ is recursively updated by
\begin{align}
\begin{split}
\label{OGA}
\hat{J}_{m}=\hat{J}_{m-1} \bigcup \{\hat{j}_m\},
\end{split}
\end{align}
where
\begin{align*}
\hat{j}_m= \arg \max_{1\leq j \leq p, j \notin \hat{J}_{m-1}} |\hat{\bm{\mu}}_{\hat{J}_{m-1}, j}|,
\end{align*}
with
$\hat{\bm{\mu}}_{J, i}=\mathbf{Z}^{\top}_{i}(\mathbf{I}-\mathbf{H}_{J})\mathbf{Y}/(n^{1/2}\|\mathbf{Z}_i\|)$,
$\|\mathbf{a}\|$ denoting the $L_2$-norm of vector $\mathbf{a}$,
and $\mathbf{H}_{J}, J\subseteq {\cal P}\equiv\{1, \ldots, p\}$, being the orthogonal
projection matrix onto the linear span of $\{\mathbf{Z}_{i}, i \in J\}$
($\mathbf{H}_{\emptyset}=\mathbf{0}$).

To investigate the performance of OGA, we make the following distributional assumptions:
\begin{description}
\item[{\rm (A1)}]
There exists $c_1^{*}>0$ such that
\begin{align}
\label{ts1.1}
\begin{split}
P(\max_{1\leq j \leq p}|n^{-1}\sum_{i=1}^{n}z_{ij}\varepsilon_{i}|\geq c_1^{*}(\log p)^{1/2}/n^{1/2})=o(1).
\end{split}
\end{align}
\end{description}

\begin{description}
\item[{\rm (A2)}]
\label{a2}
There exists $c_2^{*}>0$ such that
\begin{align}
\label{ts2.1}
\begin{split}
P(\max_{1\leq k,l \leq p}|n^{-1}\sum_{i=1}^{n}z_{ik}z_{il}-\rho_{kl}|\geq c_2^{*}(\log p)^{1/2}/n^{1/2})=o(1),
\end{split}
\end{align}
where $\rho_{kl}=\mathrm{E}(z_{1k}z_{1l})$.
\end{description}
The following examples help illustrate (A1) and (A2).
Let $\lambda_{\min}(A)$ ($\lambda_{\max}(A)$)
denote the minimum (maximum) eigenvalue of matrix $A$
and $\|\mathbf{a}\|_1$ the $L_1$-norm of vector $\mathbf{a}$.
\vspace{0.1cm}

\noindent
{\bf Example 1} (Gaussian linear processes.)
Let
\begin{eqnarray}
\label{aug2}
x_{tl}=\sum_{j=0}^{\infty} \bm{w}_{j}^{\top}(l) \bm{\delta}_{t-j},\,\,
\varepsilon_t=\sum_{j=0}^{\infty} \bm{w}_{j}^{\top}(0) \bm{\delta}_{t-j},
\end{eqnarray}
where $\bm{\delta}_{t}=(\delta_{t1}, \ldots \delta_{tq})^{\top}$
are i.i.d. Gaussian random vectors satisfying
\begin{align}
\label{nov1}
\mathrm{E}(\bm{\delta}_{t})=\mathbf{0}, \,\, \max_{1\leq i \leq q}\mathrm{E}(\delta_{ti}^2)<\bar{c}<\infty, \,\,
\lambda_{\min}(\mathrm{E}(\bm{\delta}_{t}\bm{\delta}^{\top}_{t}))>\underline{c}_0>0,
\end{align}
and $\bm{w}_{j}(l)$ obey
\begin{align}
\label{aug3}
\max_{0\leq l \leq p} \sum_{j=0}^{\infty} \|\bm{w}_{j}(l)\|_1 <M_{2}<\infty,\quad
\min_{0\leq l \leq p} \|\bm{w}_{0}(l)\|>\underline{c}_1>0.
\end{align}
Then, by making use of the Hanson-Wright inequality (see Theorem 1.1 of
Rudelson and Vershynin (2013)), it is shown in Section S1 of the supplementary material
that
\begin{align}
\label{rev32}
\mbox{(A1) and (A2) hold true under \eqref{aug2}--\eqref{aug3} and \eqref{aug5}},
\end{align}
where \eqref{aug5} is given by
\begin{eqnarray}
\label{aug5}
p\to \infty \,\, \mbox{as}\,\, n\to \infty, \,\,\frac{\log p}{n}=o(1).
\end{eqnarray}
As an application, we consider
a high-dimensional ARX model,
\begin{eqnarray}
\label{aug12}
y_t=\sum_{j=1}^{q_0}a_{j}y_{t-j}+\sum_{l=1}^{p_1}\sum_{j=1}^{r_l}
\beta_{j}^{(l)}x_{t-j+1}^{(l)}+\varepsilon_t,
\end{eqnarray}
in which $p=q_0+\sum_{l=1}^{p_1}r_l$
satisfies \eqref{aug5},
$1-\sum_{j=1}^{q_0}a_jz^{j} \neq 0$
for all $|z|\leq 1+\iota$ with $\iota$ being some positive constant,
$\sum_{j=1}^{q_0}|a_j|+\sum_{l=1}^{p_1}\sum_{j=1}^{r_l}
|\beta_{j}^{(l)}|<M_5<\infty$, $x_{t}^{(l)}=\epsilon_{t}^{(l)}+
\sum_{j=1}^{\infty}b_j^{(l)}\epsilon_{t-j}^{(l)}$, with
$\sum_{j=1}^{\infty}|b_j^{(l)}|<M_6<\infty$
for all $1\leq l\leq p$, and $\bm{\delta}_t=(\epsilon_{t}^{(1)},\ldots,\epsilon_{t}^{(p)},\varepsilon_t)^{\top}$
are i.i.d. $(p+1)$-dimensional Gaussian random vectors
obeying \eqref{nov1} with $q=p+1$.
It is not difficult to see that \eqref{aug2} and \eqref{aug3} are fulfilled by
the regressor variables and the error term in \eqref{aug12}.
Hence (A1) and (A2) are applicable to model \eqref{aug12}.

\vspace{0.2cm}
\noindent
{\bf Example 2} (Linear processes with sub-Gaussian innovations.)
Suppose that \eqref{aug2}--\eqref{aug3} and \eqref{aug5} are satisfied except that
the Gaussianity of $\bm{\delta}_{t}$ is replaced by
\begin{equation}\label{rev31a}
\lVert\delta_{tk}\rVert_{\psi_{2}}\leq L, \quad k=1,\ldots, q,
\end{equation}
where $\lVert\cdot\rVert_{\psi_{2}}$ denotes
the $\psi_{2}$ Orlicz norm and
$L$ is some positive number.
We note that \eqref{rev31a}
is fulfilled by sub-Gaussian random variables.
Assume $q= p^{s}$ for some $0\leq s<\infty$. Then by making use of the concentration inequality
given in Theorem 1.4 of Adamczak and Wolff (2015), it can be shown that (A1) and (A2) hold for some large $c_1^*$ and $c_2^{*}$.
For more details, see Huang and Ing (2019).
In addition, the regressor variables and the error term in
\eqref{aug12}
still obey (A1) and (A2), provided
assumption \eqref{rev31a}
is used in place of the
Gaussian assumption in Example 1.

\vspace{0.1cm}
We also need a sparsity condition on regression coefficients:
\begin{description}
\item[{\rm (A3)}]
\label{a3}
There is $0<\bar{M}_0<\infty$
such that $\sum_{j=1}^{p}  \beta_{j}^{*^2}\leq \bar{M}_0$.
In addition,
there exist $\gamma \geq 1$ and $0< C_{\gamma}< \infty$ such
that for any $J \subseteq {\cal P}$,
\begin{eqnarray}
\label{ts3}
\sum_{j \in J}|\beta_{j}^{*}|\leq C_{\gamma} (\sum_{j \in J} \beta_{j}^{*^2})^{(\gamma-1)/(2\gamma-1)}.
\end{eqnarray}
\end{description}
When
$\gamma=1$, \eqref{ts3} and \eqref{ts4} are equivalent.
However, \eqref{ts3} is weaker than \eqref{ts4} for $\gamma>1$.
To see this, note that if \eqref{ts4} is true for some $\gamma>1$,
then by H\"{o}lder's inequality,
\begin{align*}
\begin{split}
&\sum_{j \in J}|\beta_{j}^{*}| \leq
(\sum_{j \in J}|\beta_{j}^{*}|^{1/\gamma})^{\gamma/(2\gamma-1)}
(\sum_{j \in J} \beta_{j}^{*^{2}})^{(\gamma-1)/(2\gamma-1)}\\
& \leq M_4^{\gamma/(2\gamma-1)}
(\sum_{j \in J} \beta_{j}^{*^{2}})^{(\gamma-1)/(2\gamma-1)},
\end{split}
\end{align*}
implying that \eqref{ts3}
holds for $C_\gamma= M_4^{\gamma/(2\gamma-1)}$.
In view of the connection between
\eqref{ts3} and \eqref{ts4}, the parameter $\gamma$ in \eqref{ts3} can be understood as an index to describe
the degree of sparseness in the underlying high-dimensional models.
The larger the $\gamma$ is, the sparser the model is.
Although assumptions similar to \eqref{ts4}
are quite popular for high-dimensional regression analysis (see, e.g., Wang et al. (2014)),
there is a subtle difference between \eqref{ts3} and \eqref{ts4}.
To see this, assume that
\eqref{ts5} holds for some $\gamma>1$.
Then,
\eqref{ts3} holds for the same $\gamma$ (see Lemma \ref{l2} in the Appendix),
whereas \eqref{ts4} is violated 
due to
$L (1 +\log p) \leq \sum_{j=1}^{p}|\beta^{*}_{j}|^{1/\gamma} \leq U(1+\log p)$.
It is worth mentioning that
\eqref{ts5} not only plays an important role in time series modeling,
it also allows us to demonstrate that
the approximation error of the population counterpart of OGA
(which is
defined at the beginning of Section \ref{sec2.2} and is referred to as the population OGA)
is almost as small as that of the best $m$-term approximation
(see \eqref{may19} and Lemma \ref{l3} in the Appendix).
In the sequel, we refer to \eqref{ts3} as the `polynomial decay' case,
owing to its connection with \eqref{ts5}.
To broaden OGA's applications, we also consider a coefficient condition sparser than
\eqref{ts3}:
\begin{description}
\item[{\rm (A4)}]
There exists $0<M_0<\infty$ such that
$\max_{1\leq j\leq p}|\beta_{j}^{*}|\leq M_0$.
Moreover, there exists $M_{1} > 1$ such that for any $J \subseteq {\cal P}$,
\begin{eqnarray}
\label{ts6}
\sum_{j \in J}|\beta^{*}_{j}|\leq M_{1} \max_{j \in J} |\beta^{*}_{j}|.
\end{eqnarray}
\end{description}
Assumption (A4) is referred to as the `exponential decay' case
because \eqref{intro3} is included by \eqref{ts6}.

The following assumption on the covariance structure of
$\mathbf{z}_t=(z_1,\ldots, z_p)^{\top}$
is frequently used throughout the paper.
Define
$\bm{\Gamma}(J)=\mathrm{E}\{\mathbf{z}_t(J)\mathbf{z}_t^{\top}(J)\}$ and
$\bm{g}_{i}(J)=\mathrm{E}(z_{ti}\mathbf{z}_t(J))$,
where $J \subseteq {\cal P}$ and
$\mathbf{z}_t(J)=(z_{ti}, i\in J)^{\top}$.

\begin{description}
\item[{\rm (A5)}]
\label{a5}
For some positive numbers $\bar{D}$ and $M$,
\begin{align}
\label{ts8}
 \max_{1 \leq \sharp(J) \leq \bar{D}(n/\log p)^{1/2}, i \notin J}\|\bm{\Gamma}^{-1}(J)
\bm{g}_{i}(J)\|_{1} < M,
\end{align}
where $\sharp(J)$ denotes the cardinality of $J$ .
\end{description}
Since
$\bm{\Gamma}^{-1}(J)
\bm{g}_{i}(J)=\arg \min_{\bm{c}\in R^{\sharp(J)}} \mathrm{E}(z_{ti}-\bm{c}^{\top}\mathbf{z}_t(J))^{2}$,
\eqref{ts8} essentially says that the regression coefficients for
$z_{ti}$ on $\mathbf{z}_t(J)$ with all $i \notin J $
and $\sharp(J)\leq \bar{D}(n/\log p)$ are $L_1$ bounded.
This condition holds even when $z_{t1}, \cdots, z_{tp}$ are highly correlated;
see Section S3 of the supplementary document.
Let $\bm{g}_{y}(J)=\mathrm{E}(y_t\mathbf{z}_t(J))$ and
$\bm{\beta}^{*}(J)=\bm{\Gamma}^{-1}(J)\bm{g}_{y}(J)$$=\arg \min_{\bm{c}\in R^{\sharp(J)}}\mathrm{ E}(y_t-\bm{c}^{\top}\mathbf{z}_t(J))^{2}$,
which is the regression coefficients for $y_t$ on $\mathbf{z}_{t}(J)$.
By making use of \eqref{ts8}, we will show later that for any $J \subseteq{\cal P}$
with $\sharp(J)\leq \bar{D}(n/\log p)^{1/2}$, there exists $0<C<\infty$ such that
\begin{eqnarray}
\label{ubc}
\|\bm{\beta}^{*}-\bm{\beta}^{*}(J)\|_1 \leq C \sum_{j \notin J} |\beta^{*}_{j}|,
\end{eqnarray}
where
$\bm{\beta}^{*}=(\beta_1^{*}, \ldots, \beta^*_{p})^{\top}$ and
$\bm{\beta}^{*}(J)$ here is regarded as a $p$-dimensional vector
with undefined entries set to 0.
Inequality \eqref{ubc} is referred to as the uniform Baxter's inequality.
(For more details on Baxter's inequality in autoregressive modeling, see Baxter (1962), Berk (1974), and Pourahmadi (1989).)
This inequality can be used together with
\eqref{ts3} to yield, for all $\sharp(J) \leq \bar{D}(n/\log p)^{1/2}$,
\begin{eqnarray}
\label{ubc1}
\|\bm{\beta}^{*}-\bm{\beta}^{*}(J)\|_1 \leq C C_\gamma
(\sum_{j \notin J}\beta^{*^2}_{j})^{(\gamma-1)/(2\gamma-1)},
\end{eqnarray}
which is one of the key ingredients in our asymptotic analysis of OGA+HDAIC.

To derive \eqref{ubc} from \eqref{ts8},
we may assume without loss of generality that
$J=\{j_1, \ldots, j_{q}\}$ for some $1\leq q \leq \bar{D}(n/\log p)^{1/2}$,
where $j_i, i=1, \ldots q$,
are distinct elements in ${\cal P}$.
Note first that
\begin{eqnarray}
\label{ubc1}
\|\bm{\beta}^{*}-\bm{\beta}^{*}(J)\|_1 \leq \|\bm{\beta}^{*}(J)-\bm{\beta}^{*}_J\|_1+
\sum_{j \notin J} |\beta^{*}_{j}|,
\end{eqnarray}
where $\bm{\beta}^{*}_J=(\beta^{*}_{j_1}, \ldots, \beta^{*}_{j_q})^{\top}$.
Denote $\bm{g}_{y}(J)$ by $(\gamma_{y,j_1}, \ldots, \gamma_{y, j_q})^{\top}$.
Then, it follows that $\gamma_{y,j_i}=\sum_{l=1}^{p}\rho_{j_il}\beta_{l}^{*}$, $i=1, \ldots, q$,
and hence
\begin{align*}
\begin{split}
&\bm{\Gamma}(J)^{-1}\bm{\Gamma}(J)(\bm{\beta}^{*}(J)-\bm{\beta}^{*}_J)=
\bm{\Gamma}(J)^{-1}(\sum_{l \notin J}\rho_{j_1l}\beta_l^{*}, \ldots, \sum_{l \notin J}\rho_{j_ql}\beta_l^{*})^{\top}\\
&=\sum_{l \notin J}\beta_{l}^{*}\bm{\Gamma}(J)^{-1}\bm{g}_{l}(J).
\end{split}
\end{align*}
Taking the $L_1$-norm on both sides, \eqref{ubc} (with $C=M+1$) follows from
\eqref{ubc1} and \eqref{ts8}.

Before closing this section,
we remark that
\eqref{ts6} can be viewed as a limiting case of \eqref{ts3}.
To see this,
note that \eqref{ts6} implies
that
for any $J \subseteq {\cal P}$,
$\sum_{j \in J}|\beta_{j}^{*}|\leq M_1(\sum_{j \in J}\beta_{j}^{*^2})^{\lim_{\gamma \to \infty}(\gamma-1)/(2\gamma-1)}$.
In addition, the strong sparsity condition,
\begin{eqnarray}
\label{aug13}
k_0=\sharp(N_n) <M_7,
\end{eqnarray}
where $N_n=\{j: \beta_{j}^{*} \neq 0, 1\leq j \leq p \}$
and $M_7$ is some positive integer,
is also a limiting case of \eqref{ts3} because
\eqref{aug13} yields that
for any $J \subseteq {\cal P}$,
$\sum_{j \in J}|\beta_{j}^{*}|\leq M_7^{1/2}(\sum_{j \in J}\beta_{j}^{*^2})^{\lim_{\gamma \to \infty}(\gamma-1)/(2\gamma-1)}$.

\subsection{Rates of Convergence of the OGA}
\label{sec2.3}
Let $\mathbf{x}=(x_1, \ldots x_p)^{\top}$
be independent of and have the same covariance structure as $\{\mathbf{x}_t\}$
and $y(\mathbf{x})=\sum_{j=1}^{p}\beta_jx_j$.
Then, $y(\mathbf{x})$ can be predicted by
$\hat{y}_m(\mathbf{x})=\mathbf{x}^{\top}(\hat{J}_m) \hat{\bm{\beta}}(\hat{J}_m)$,
where $\mathbf{x}(J)=(x_i, i \in J)$
and $\hat{\bm{\beta}}(J)=(\sum_{t=1}^{n}\mathbf{x}_t(J)\mathbf{x}_t^{\top}(J))^{-1}\sum_{t=1}^{n}\mathbf{x}_t(J)y_t$.
Note also that
$\hat{y}_m(\mathbf{x})=\mathbf{z}^{\top}(\hat{J}_m)\hat{\bm{\beta}}^{*}(\hat{J}_m)$,
where
$\mathbf{z}(J)=(z_i, i \in J)$ with $z_i=x_i/\sigma_i$, and
$\hat{\bm{\beta}}^{*}(\hat{J}_m)=(\sum_{t=1}^{n}\mathbf{z}_t(J)\mathbf{z}_t^{\top}(J))^{-1}\sum_{t=1}^{n}\mathbf{z}_t(J)y_t$.
One of the most natural performance measures for
$\hat{y}_m(\mathbf{x})$ is
the conditional mean squared prediction error (CMSPE),
\begin{align}
\begin{split}
\label{aug14}
&\mathrm{E}_n\{y(\mathbf{x})-\hat{y}_{m}(\mathbf{x})\}^{2}=
\mathrm{E}_n\{y(\mathbf{x})-y_{\hat{J}_m}(\mathbf{x})\}^{2}+
\mathrm{E}_n\{y_{\hat{J}_m}(\mathbf{x})-\hat{y}_{m}(\mathbf{x})\}^{2},
\end{split}
\end{align}
where
$\mathrm{E}_n(\cdot)=\mathrm{E}(\cdot|y_{1}, \mathbf{x}_{1},
\cdots,y_{n}, \mathbf{x}_{n})$,
and
$y_{J}(\mathbf{x})$$=
\bm{\beta}^{*^{\top}}(J)\mathbf{z}(J)$.
A convergence rate of the left-hand side of \eqref{aug14} is established in the next theorem.
\begin{thm}
\label{t1}
Suppose that \eqref{aug1}, {\rm (A1)--(A3)}, {\rm (A5)},
\begin{align}
\label{ts10}
\lambda_{\min}(\boldG)\geq \lambda_{1}>0,
\end{align}
and
\begin{eqnarray}
\label{kn}
\log p=o(n), \,\,K_{n}=\bar{\delta}\left(\frac{n}{\log p}\right)^{1/2}
\end{eqnarray}
hold, where
$\boldG= \mathrm{E}(\mathbf{z} \mathbf{z}^{\top})$ and
$0<\bar{\delta}<\min\{\bar{\tau}, \bar{D}\}$, with $\bar{D}$
defined in assumption {\rm (A5)} and
\begin{align}
\label{sep7}
\bar{\tau}=\sup \bm{\tau} \equiv \sup\{\tau: \tau>0,\,\,
\limsup_{n \to \infty}\frac{\tau c_2^{*}}
{\displaystyle{\min_{\sharp(J)\leq \tau (n/\log p)^{1/2}}}\lambda_{\min}(\bm{\Gamma}(J))}\leq 1\}.
\end{align}
Then,
\begin{eqnarray}
\label{ts20}
&& \max_{1 \leq m \leq K_{n}}\left(\frac{\mathrm{E}_n
\{y(\mathbf{x})-\hat{y}_{m}(\mathbf{x})\}^{2}}{m^{-2\gamma+1}+n^{-1}m \log p}\right)
= O_{p}(1).
\end{eqnarray}
Moreover, if {\rm (A4)} holds instead of {\rm (A3)},
then
\begin{align}
\label{ts21}
&& \max_{1 \leq m \leq K_{n}}\left(\frac{\mathrm{E}_n
\{y(\mathbf{x})-\hat{y}_{m}(\mathbf{x})\}^{2}}{{\rm exp}(-G_{3}m)+n^{-1}m \log p}\right)
= O_{p}(1),
\end{align}
where
$G_{3}$ is some positive constant given in \eqref{ts12} in the Appendix.
\end{thm}

\noindent
{\bf Proof.}
We first prove \eqref{ts20}.
Recall $\hat{J}_{k}=\{\hat{j}_{1}, \cdots, \hat{j}_{k}\}$ and define
\begin{eqnarray*}
\hat{\mu}_{J, i}=
\frac{n^{-1}\sum_{t=1}^{n}(y_{t}-\hat{y}_{t; J})x_{ti}}{(n^{-1}\sum_{t=1}^{n}x^{2}_{ti})^{1/2}}=
\frac{n^{-1}\sum_{t=1}^{n}(y_{t}-\hat{y}_{t; J})z_{ti}}{(n^{-1}\sum_{t=1}^{n}z^{2}_{ti})^{1/2}},
\end{eqnarray*}
where $(\hat{y}_{1; J}, \ldots, \hat{y}_{n; J})^{\top}=\mathbf{H}_{J}\mathbf{Y}$.
Moreover, let
\begin{eqnarray*}
A_{n}(m)\!=\!\left\{ \max_{(J, i): \sharp(J)\leq m-1, i \notin J}
|\hat{\mu}_{J, i}-\mu_{J,i}|
\leq s (\log p/n)^{1/2} \right\},
\end{eqnarray*}
and
\begin{eqnarray*}
B_{n}(m)=\left\{ \min_{0\leq i \leq m-1} \, \max_{1\leq j
\leq p}
|\mu_{\hat{J}_{i}, j}|> \tilde{\xi}s (\log p/n)^{1/2}
\right\},
\end{eqnarray*}
where
$\mu_{J, i}=\mathrm{E}[(y(\mathbf{x})-y_{J}(\mathbf{x}))z_{i}]$,
$s>0$ is some large constant, and
$\tilde{\xi}=2/(1-\xi)$ with $0< \xi <1$ being arbitrarily given.

By an argument similar to that of (3.10) in Ing and Lai (2011), it follows that
for all $1 \leq q \leq m$,
\begin{eqnarray*}
\label{ts22}
|\mu_{\hat{J}_{q-1}, \hat{j}_{q}}| \geq \xi
\max_{1\leq i \leq p}
|\mu_{\hat{J}_{q-1}, i}| \,\,\mbox{on} \,\, A_{n}(m) \bigcap B_{n}(m).
\end{eqnarray*}
This and \eqref{ts11} in the Appendix,
which gives an error bound for
the population OGA under (A3),
lead to
\begin{align}
\label{ts23}
\begin{split}
\mathrm{E}_{n}(y(\mathbf{x})-y_{\hat{J}_{m}}(\mathbf{x}))^{2}
\leq G_{1}m^{-2\gamma+1} \,\,\mbox{on} \,\, A_{n}(m) \bigcap B_{n}(m).
\end{split}
\end{align}
Moreover, (A3) and \eqref{ts10} imply that for any $0\leq i \leq m-1$,
\begin{align*}
\begin{split}
& \mathrm{E}_{n}(y(\mathbf{x})-y_{\hat{J}_{i}}(\mathbf{x}))^{2} \leq \max_{1 \leq j \leq p}|\mu_{\hat{J}_{i}, j}|
\sum_{l=1, l \notin \hat{J}_{i}}^{p}|\beta^*_{l}| \\
&\leq C_{\gamma} \max_{1 \leq j \leq p}|\mu_{\hat{J}_{i}, j}|
 (\sum_{l=1, l \notin \hat{J}_{i}}^{p}\beta^{*^2}_{l})^{(\gamma-1)/(2\gamma-1)} \\
&\leq C_{\gamma} \max_{1 \leq j \leq p}|\mu_{\hat{J}_{i}, j}| \lambda^{-(\gamma-1)/(2\gamma-1)}_{1}
 (\mathrm{E}_{n}(y(\mathbf{x})-y_{\hat{J}_{i}}(\mathbf{x}))^{2})^{(\gamma-1)/(2\gamma-1)},
\end{split}
\end{align*}
and hence
\begin{align}
\label{ts24}
\mathrm{E}_{n}(y(\mathbf{x})-y_{\hat{J}_{i}}(\mathbf{x}))^{2}
\leq (C_{\gamma} \max_{1 \leq j \leq p}|\mu_{\hat{J}_{i}, j}|)^{2-\gamma^{-1}}
\lambda^{-1+\gamma^{-1}}_{1}.
\end{align}
By \eqref{ts24},
\begin{align}
\label{ts25}
\begin{split}
& \mathrm{E}_{n}(y(\mathbf{x})-y_{\hat{J}_{m}}(\mathbf{x}))^{2} \leq \min_{0 \leq i \leq m-1}
\mathrm{E}_{n}(y(\mathbf{x})-y_{\hat{J}_{i}}(\mathbf{x}))^{2} \\
&\leq
C^{2-\gamma^{-1}}_{\gamma}\lambda^{-1+\gamma^{-1}}_{1}
(\min_{0\leq i \leq m-1} \max_{1 \leq j \leq p}|\mu_{\hat{J}_{i}, j}|)^{2-\gamma^{-1}} \\
&\leq
C^{2-\gamma^{-1}}_{\gamma}\lambda^{-1+\gamma^{-1}}_{1}
(\tilde{\xi}s)^{2-\gamma^{-1}}
(n^{-1}\log p)^{1-(2\gamma)^{-1}} \,\,\mbox{on}\,\,
B^{c}_{n}(m).
\end{split}
\end{align}
Since $A_{n}(m)$ decreases as $m$ increases, \eqref{ts23} and \eqref{ts25} yield that
for all $1 \leq m \leq K_{n}$ and some $C_{2}>0$,
\begin{align}
\label{ts26}
\begin{split}
\mathrm{E}_{n}(y(\mathbf{x})-y_{\hat{J}_{m}}(\mathbf{x}))^{2}I_{A_{n}(K_{n})} \leq
C_{2} \max \{m^{-2\gamma+1}, \{n^{-1} \log p\}^{1-(2\gamma)^{-1}}\}.
\end{split}
\end{align}

We show
in Section S1 of the supplementary material that
\begin{eqnarray}
\label{332}
P (\|\hat{\bm{\Gamma}}^{-1}(\hat{J}_{K_n})\| \leq \bar{B})=1+o(1),
\end{eqnarray}
where
$\hat{\bm{\Gamma}}(J)=n^{-1}\sum_{t=1}^{n}\mathbf{z}_{t}(J)\mathbf{z}^{\top}_{t}(J)$
and
\begin{align}
\label{sep8}
\bar{B}>\frac{1}{\displaystyle{\liminf_{n \to \infty}}
\displaystyle{\min_{\sharp{(J)}\leq K_n }}\lambda_{\min}(\bm{\Gamma}(J))-c_2^{*}\bar{\delta}},
\end{align}
noting that the positiveness of the denominator is ensured by \eqref{kn} and \eqref{sep7}.
With the help of \eqref{332},
(A1), (A2), and (A5), it is shown in the same section that
there exists a sufficiently large $s$ such that
\begin{eqnarray}
\label{ts-1}
\lim_{n \to \infty} P(A_{n}(K_{n}))=1,
\end{eqnarray}
which, together with \eqref{ts26}, yields
\begin{align}
\label{ts27}
\max_{1\leq m \leq K_{n}}\frac{\mathrm{E}_{n}(y(\mathbf{x})-y_{\hat{J}_{m}}(\mathbf{x}))^{2}}
{\max \{m^{-2\gamma+1}, (n^{-1} \log p)^{1-(2\gamma)^{-1}}\}}=
O_{p}( 1).
\end{align}
Moreover, we have
\begin{eqnarray}
\label{ts28}
\max_{1 \leq m \leq K_{n}}
\frac{n\mathrm{E}_{n}[\{\hat{y}_{m}(\mathbf{x})-y_{\hat{J}_{m}}(\mathbf{x})\}^{2}]}{m \log p}=O_{p}(1),
\end{eqnarray}
which is also proved in Section S1 of the supplementary material.
In view of \eqref{ts27}, \eqref{ts28}, and
the fact that $(\log p/n)^{1-(2\gamma)^{-1}} \leq m^{-2\gamma+1}$
if $m \leq (n/\log p)^{(2\gamma)^{-1}}$
and $(\log p/n)^{1-(2\gamma)^{-1}} \leq n^{-1}m \log p$ if
$m \geq (n/\log p)^{(2\gamma)^{-1}}$,
the desired conclusion \eqref{ts20} follows.
Equation \eqref{ts21} follows from
\eqref{ts12} in the Appendix (which gives an error bound for
the population OGA under (A4)) and
an argument similar to that used to prove \eqref{ts20}.
We skip the details in order to save space.
\hfill$\Box$

\vspace{0.1cm}
\noindent
{\bf Remark 1.}
It is easy to see that the $\bm{\tau}$ defined in \eqref{sep7} is nonempty.
In particular, $x \in \bm{\tau}$ for any $x \in (0, \lambda_1/c_2^{*}]$.
It is also not difficult to see that $\bar{\tau}< a/c_2^{*}$ for any $a>1$.

\vspace{0.1cm}
In view of \eqref{ts20}, to strike
a suitable balance between squared bias and variance,
one should choose $m \approx (n/ \log p)^{1/2\gamma}$
in the polynomial decay case, which yields
a rate of convergence,
$(n^{-1} \log p)^{1-(2\gamma)^{-1}}$.
Similarly, \eqref{ts21} suggests that
the best convergence rate one can expect
in the exponential decay case is
$n^{-1} \log n \log p$, which is ensured
by selecting $m \approx \log n/G_3$.
The optimality of the rates,
$(n^{-1} \log p)^{1-(2\gamma)^{-1}}$ and
$n^{-1} \log n \log p$,
will be discussed further in Section \ref{sec3.11}.
In most practical situations, however,
not only do we not know what $\gamma$ or $G_3$ is,
we do not even know which
of (A3) and (A4) is true.
To attain the aforementioned optimal convergence rates
without knowing the degree of sparseness,
a data-driven method to determine the number of OGA iterations
is called for.
In the next section, we show that
HDAIC (see \eqref{hdic}) can
fulfill this need.

Finally, we note that
if \eqref{ts1.1} and \eqref{ts2.1} are weakened to
\begin{align}
\label{sep1}
\begin{split}
P(\max_{1\leq j \leq p}|n^{-1}\sum_{i=1}^{n}z_{ij}\varepsilon_{j}|\geq c_1^{*}(\log p)^{(1+\bar{c}_1)/2}/n^{1/2})=o(1),
\end{split}
\end{align}
and
\begin{align}
\label{sep2}
\begin{split}
P(\max_{1\leq k,l \leq p}|n^{-1}\sum_{i=1}^{n}z_{ik}z_{il}-\rho_{kl}|\geq c_2^{*}(\log p)^{(1+\bar{c}_2)/2}/n^{1/2})=o(1),
\end{split}
\end{align}
respectively, where $0\leq \bar{c}_1, \bar{c}_2<\infty$
are some constants, and \eqref{kn} is strengthened to
\begin{eqnarray}
\label{kn1}
(\log p)^{1+\bar{c}}=o(n)\,\,\mbox{and}\,\,K_{n}=\underline{\delta}\frac{n^{1/2}}{(\log p)^{(1+\bar{c})/2}},
\end{eqnarray}
where $\bar{c}=\max\{\bar{c}_1,\bar{c}_2\}$
and $\underline{\delta}$ is some positive constant,
then \eqref{ts20} and \eqref{ts21} become
\begin{eqnarray}
\label{sep3}
&& \max_{1 \leq m \leq K_{n}}\left(\frac{\mathrm{E}_n
\{y(\mathbf{x})-\hat{y}_{m}(\mathbf{x})\}^{2}}{m^{-2\gamma+1}+n^{-1}m (\log p)^{1+\bar{c}}}\right)
= O_{p}(1),
\end{eqnarray}
and
\begin{align}
\label{sep4}
&& \max_{1 \leq m \leq K_{n}}\left(\frac{\mathrm{E}_n
\{y(\mathbf{x})-\hat{y}_{m}(\mathbf{x})\}^{2}}{{\rm exp}(-G_{3}m)+n^{-1}m (\log p)^{1+\bar{c}}}\right)
= O_{p}(1),
\end{align}
respectively.
While \eqref{sep1} and \eqref{sep2}
are satisfied by a broader class of time series models (see Wu and Wu (2016) for a detailed discussion),
to determine the optimal $m$ in \eqref{sep3} or \eqref{sep4},
the HDAIC must also be corrected according to the value of $\bar{c}$.
This kind of correction, however, is hardly implemented in practice because $\bar{c}$ is in general unknown.


\section{Analysis of OGA+HDAIC}
\label{sec3}
In Section \ref{sec3.1}, the rate of convergence of OGA+HDAIC is
established under various sparsity conditions; see Theorem \ref{thm3.1}.
Comparisons of Theorem \ref{sec3.1} and related existing results are given in Section \ref{sec3.11}.
The proof of Theorem \ref{thm3.1} is provided in Section \ref{sec3.2}.

\subsection{Error bounds for OGA+HDAIC}
\label{sec3.1}
\indent Define
\begin{eqnarray}
\label{hdic}
{\rm HDAIC}(J)=\left(1+\frac{s_a \sharp(J) \log p}{n}\right)\hat{\sigma}_J^2,
\end{eqnarray}
where
$\hat{\sigma}_J^2=n^{-1}\mathbf{Y}^{\top}(\mathbf{I}-\mathbf{H}_{J})\mathbf{Y}$
and $s_a$ is some positive constant, and define
\begin{eqnarray*}
\hat{k}_{n}=\arg\min_{1\leq k\leq K_n}{\rm HDAIC}(\hat{J}_k),
\end{eqnarray*}
noting that $\hat{J}_k$ is defined in \eqref{OGA}.

\begin{thm}
\label{thm3.1}
Suppose that \eqref{aug1}, {\rm (A1), (A2), (A5)},
\eqref{ts10}, \eqref{kn}, and
\begin{eqnarray}
\label{3-0}
n^{-1}\sum_{t=1}^{n}\varepsilon_t^2=\sigma^2+o_p(1)
\end{eqnarray}
hold.
Then, for
\begin{eqnarray}
\label{hdic1}
s_a>\bar{V}_0\equiv\frac{2\bar{B}(c_1^{*^2}+c_2^{*^2})}{\sigma^{2}},
\end{eqnarray}
where
$\bar{B}$ is defined in \eqref{sep8}, we have
\begin{description}
  \item[(i)]
\begin{eqnarray}
\label{3-3}
\frac{\mathrm{E}_n\left(y(\mathbf{x})-\hat{y}_{\hat{k}_{n}}(\mathbf{x})\right)^2}
{\left( \frac{\log p}{n}\right)^{1-1/2\gamma}}=O_p(1),
\end{eqnarray}
provided {\rm (A3)} is true,

  \item[(ii)]
\begin{eqnarray}
\label{3-2}
\frac{\mathrm{E}_n\left(y(\mathbf{x})-\hat{y}_{\hat{k}_{n}}(\mathbf{x})\right)^2}{\frac{\log n\log p}{n}}=O_p(1),
\end{eqnarray}
provided {\rm (A4)} is true and $\log p= o(n/(\log n)^{2})$,

  \item[(iii)]
\begin{eqnarray}
\label{3-1}
\frac{\mathrm{E}_n\left(y(\mathbf{x})-\hat{y}_{\hat{k}_{n}}(\mathbf{x})\right)^2}{\frac{k_0\log p}{n}}=O_p(1),
\end{eqnarray}
provided $\mathrm{E}(y_{t}^{2})$ is bounded above by a finite constant and
\begin{align}
\label{sep6}
\begin{split}
& \min_{j \in N_n}|\beta_j^{*}|\geq \underline{\theta}, \,\,\mbox{for some}\,\, \underline{\theta}>0,\\
& k_0 (\sum_{j \in N_n}|\beta_j^{*}|)^{2}  = o(n/\log p).
\end{split}
  \end{align}
\end{description}
\end{thm}

\vspace{0.1cm}
\noindent
{\bf Remark 2.} The sparsity condition
\eqref{sep6} implies
$k_0=o\big((n/\log p)^{1/3}\big)$, allowing
$k_0$ to grow to $\infty$ slowly with $n$.
Moreover, \eqref{3-1} also holds when \eqref{ts10} is weakened to
\begin{align}
  \label{sep55}
      \min_{\sharp(J) \leq \eta(n/\log p)^{1/2}}\lambda_{\min}(\bm{\Gamma}(J)) \geq \lambda_1,
  \end{align}
for some $\eta>0$; see Section S2 in the supplementary document.
However, since it is unknown which kind of sparsity condition is true
among those described in (i), (ii), and (iii) of Theorem \ref{thm3.1},
and since \eqref{ts10} appears to be indispensable for the proofs of
\eqref{3-3} and \eqref{3-2}, the latter assumption is still adopted in our unified theory.

\vspace{0.1cm}
\noindent
{\bf Remark 3.}
We briefly discuss extensions of Theorems \ref{t1}
and \ref{thm3.1} to the following multivariate time series models,
\begin{eqnarray}
\label{rev24a}
\bm{y}_t=\sum_{l=1}^{p} \bm{b}_{j}x_{tj}+\bm{e}_t, \,\,t=1,\ldots,n,
\end{eqnarray}
where $\bm{y}_t$, $\bm{e}_t$,
and $\bm{b}_{j}$ are $d$-dimensional vectors, $d$ is allowed to grow to infinity with $n$, and
$\{(\bm{e}^{\top}_t, \mathbf{x}^{\top}_{t})^{\top}\}$
is a zero-mean stationary time series
satisfying $\mathrm{E}(\mathbf{x}_{t}\bm{e}^{\top}_t)=\mathbf{0}$.
Define
$\hat{\bm{\psi}}_{J, i}=\|{\cal Y}^{\top}(\mathbf{I}-\mathbf{H}_{J})\mathbf{Z}^{\top}_{i}\|/(n^{1/2}\|\mathbf{Z}_i\|)$,
where ${\cal Y}=(\bm{y}_1, \cdots, \bm{y}_n)^{\top}$, and
$\mathbf{H}_{J}$ and $\mathbf{Z}_{i}$ are defined as in Section \ref{sec2.1}.
A multivariate version of OGA, MOGA, is
initialized by $\hat{L}_0=\emptyset$.
For $m \geq 1$,
$\hat{L}_{m}$ is recursively updated by
\begin{align*}
\begin{split}
\hat{L}_{m}=\hat{L}_{m-1} \bigcup \{\hat{l}_m\},
\end{split}
\end{align*}
where
$\hat{l}_m= \arg \max_{1\leq l \leq p, l \notin \hat{L}_{m-1}} \hat{\bm{\psi}}_{\hat{L}_{m-1}, l}$.
Consider multivariate extensions of the sparsity conditions (A3) and (A4):
\begin{description}
\item[{\rm (A3$^{'}$)}]
There is $0<\bar{M}_0<\infty$ such that
\begin{eqnarray*}
d^{-1}\sum_{j=1}^{p}\|\bm{b}_{j}^{*}\|^2<\bar{M}_0.
\end{eqnarray*}
Moreover, there exist $\gamma \geq 1$ and $0< C_{\gamma}< \infty$ such
that for any $J \subseteq {\cal P}$,
\begin{eqnarray*}
\sum_{j \in J}\|\bm{b}_{j}^{*}\|/d^{1/2}\leq C_{\gamma} \{\sum_{j \in J} \|\bm{b}_j^{*}\|^{2}/d\}^{(\gamma-1)/(2\gamma-1)},
\end{eqnarray*}
where $\bm{b}_{j}^{*}=\sigma_j\bm{b}_{j}$.
\end{description}
\begin{description}
\item[{\rm (A4$^{'}$)}]
There is $0<M_0<\infty$ such that
\begin{eqnarray*}
\max_{1\leq j\leq p}\|\bm{b}_{j}^{*}\|<d^{1/2}M_0.
\end{eqnarray*}
Moreover, there exists $M_{1} > 1$ such that for any $J \subseteq {\cal P}$,
\begin{eqnarray*}
\sum_{j \in J}\|\bm{b}^{*}_{j}\|
\leq M_{1} \max_{j \in J} \|\bm{b}^{*}_{j}\|.
\end{eqnarray*}
\end{description}
Moreover, a natural generalization of (A1) under model \eqref{rev24a} is
\begin{description}
\item[{\rm (A1$^{'}$)}]
There exists $c_1^{*}>0$ such that
\begin{align*}
\begin{split}
P(\max_{1\leq j \leq p, 1\leq l \leq d}|
n^{-1}\sum_{t=1}^{n}z_{tj}\varepsilon_{tl}|\geq c_1^{*}(\log pd)^{1/2}/n^{1/2})=o(1),
\end{split}
\end{align*}
where $(\varepsilon_{t1}, \ldots, \varepsilon_{td})^{\top}=\bm{e}_t$.
\end{description}
Let $\mathbf{x}=(x_1, \ldots, x_p)^{\top}$ be defined as in Section \ref{sec2.3}
and $\bm{y}(\mathbf{x})=\sum_{j=1}^{p} \bm{b}_{j}x_{j}$.
Then, $\bm{y}(\mathbf{x})$ can be predicted by
$\hat{\bm{y}}_{m}(\mathbf{x})=\hat{\bm{B}}(\hat{L}_m)^{\top}\mathbf{x}(\hat{L}_m)$,
where
$\hat{\bm{B}}(J)=(\sum_{t=1}^{n}\mathbf{x}_t(J)\mathbf{x}_t^{\top}(J))^{-1}\sum_{t=1}^{n}\mathbf{x}_t(J)\bm{y}_t^{\top}$.
Suppose that
\begin{eqnarray}
\label{rev26c}
\log pd=o(n) \quad \mbox{and} \quad K_n=\zeta (n/\log pd)^{1/2},
\end{eqnarray}
for some $\zeta>0$.
Then, under \eqref{rev26c} and the assumptions of Theorem \ref{t1}, with
(A1), (A3), and (A4) replaced by
(A1$^{'}$), (A3$^{'}$), and (A4$^{'}$), it can be shown that
\begin{eqnarray}
\label{rev26a}
&& \max_{1 \leq m \leq K_{n}}\left(\frac{d^{-1}\mathrm{E}_n
\|\bm{y}(\mathbf{x})-\hat{\bm{y}}_{m}(\mathbf{x})\|^{2}}{m^{-2\gamma+1}+n^{-1}m \log pd}\right)
= O_{p}(1),
\end{eqnarray}
and for some $G_4>0$,
\begin{align}
\label{rev26b}
&& \max_{1 \leq m \leq K_{n}}\left(\frac{d^{-1}\mathrm{E}_n
\|\bm{y}(\mathbf{x})-\hat{\bm{y}}_{m}(\mathbf{x})\|^{2}}{{\rm exp}(-G_{4}m)+n^{-1}m \log pd}\right)
= O_{p}(1).
\end{align}
To choose a suitable number of MOGA iterations, one may consider a multivariate extension of HDAIC (MHDAIC),
\begin{eqnarray*}
\label{mhdic}
{\rm MHDAIC}(J)=\left(1+\frac{\iota_a \sharp(J) \log pd}{n}\right)\hat{\bm{\Sigma}}_J,
\end{eqnarray*}
where
$\hat{\bm{\Sigma}}_J=(nd)^{-1}tr\big({\cal Y}^{\top}(\mathbf{I}-\mathbf{H}_{J}){\cal Y}\big)$
and $\iota_a$ is some positive constant, and define
\begin{eqnarray*}
\hat{m}_{n}=\arg\min_{1\leq m\leq K_n}{\rm MHDAIC}(\hat{L}_m).
\end{eqnarray*}
We conjecture that
$d^{-1}\mathrm{E}_n
\|\bm{y}(\mathbf{x})-\hat{\bm{y}}_{\hat{m}_n}(\mathbf{x})\|^{2}$
is of order $O_p((\log pd/n)^{1-1/(2\gamma)})$,
$O_p(\log n\log pd/n)$, or
$O_p(k_0\log pd/n)$ under
(A3$^{'}$),
(A4$^{'}$), or a strong sparsity condition resembling \eqref{sep6}, respectively.
However, the rigorous proof of this result
and those of \eqref{rev26a} and \eqref{rev26b} are out of the scope of this paper,
and are left for future work.

\subsection{Some comparisons with existing results}
\label{sec3.11}
It would be interesting to compare \eqref{3-3}
with Corollary 3 of Negahban et al. (2012), which
provides an error bound for Lasso in the following high-dimensional regression model,
\begin{eqnarray}
\label{yu1}
y_t=\sum_{j=1}^{p}\beta_j^{*}x_{tj}+\epsilon_t, \,\,t=1,\ldots, n,
\end{eqnarray}
where $\{\epsilon_t\}$
is a sequence of i.i.d. $N(0, \sigma^2)$
random variables and
$\{x_{tj}\}$ are non-random constants satisfying
$n^{-1}\sum_{t=1}^{n}x_{tj}^{2} \leq 1, 1\leq j \leq p$,
and the restricted eigenvalue condition defined in (31)
of their paper.
When
\begin{eqnarray}
\label{yu2}
\sum_{j=1}^{p}|\beta_j^{*}|^{1/\gamma}\leq (n/\log p)^{1-1/(2\gamma)},
\end{eqnarray}
for some $\gamma\geq 1$,
it is shown in the corollary that
\begin{eqnarray}
\label{yu3}
\|\hat{\bm{\beta}}_{\lambda_n}-\bm{\beta}^{*}\|^{2}= O_{p}
\left(\sum_{j=1}^{p}|\beta_j^{*}|^{1/\gamma}\big(\frac{\log p}{n}\big)^{1-1/(2\gamma)}\right),
\end{eqnarray}
where $\hat{\bm{\beta}}_{\lambda_n}$
is the Lasso estimate of $\bm{\beta}^{*}$
with $\lambda_n=4\sigma(\log p/n)^{1/2}$. On the other hand,
\eqref{3-3} implies that under model \eqref{aug1},
\begin{align}
\label{yu4}
\begin{split}
&\|\hat{\bm{\beta}}_{n}(\hat{J}_{\hat{k}_n})-\bm{\beta}^{*}\|^{2}
\leq \lambda_1^{-1} \mathrm{E}_n\left(y(\mathbf{x})-\hat{y}_{\hat{k}_{n}}(\mathbf{x})\right)^2\\
&= O_{p}
\left(\big(\frac{\log p}{n}\big)^{1-1/(2\gamma)}\right).
\end{split}
\end{align}
In addition to allowing for serially correlated data,
\eqref{yu4} may lead to a faster convergence rate than \eqref{yu3}.
In particular,
the bound on the right-hand side of
\eqref{yu3} is larger than that on
the right-hand side of \eqref{yu4}
by a factor of $\log p$ as $p \to \infty$
when \eqref{ts5}, with $\gamma>1$,
and \eqref{yu2} follows.

Assuming that the $\{\mathbf{x}_t\}$ and $\{\epsilon_t\}$ in \eqref{yu1}
are generated according to independent, centered,
Gaussian stationary time series,
Proposition 3.3 of Basu and Michailidis (2015) establishes for Lasso the following bounds:
\begin{eqnarray}
\label{yu5}
\|\hat{\bm{\beta}}_{\lambda_n}-\bm{\beta}^{*}\|^{2}= O_{p}
\left(\frac{k_0\log p}{n}\right),
\end{eqnarray}
and
\begin{eqnarray}
\label{yu6}
n^{-1}\sum_{t=1}^{n}(\mathbf{x}_t^{\top}(\hat{\bm{\beta}}_{\lambda_n}-\bm{\beta}^{*}))^{2}= O_{p}
\left(\frac{k_0\log p}{n}\right),
\end{eqnarray}
where $p \to \infty$, $k_0 = O(n/\log p)$, and $\lambda_n \geq c^*(\log p/n)^{1/2}$ for some $c^*>0$.
By \eqref{3-1} and an argument used in Section S2 of the supplementary document,
it can be shown that under model \eqref{aug1},
\begin{eqnarray}
\label{yu7}
\|\hat{\bm{\beta}}(\hat{J}_{\hat{k}_n})-\bm{\beta}^{*}\|^{2}= O_{p}
\left(\frac{k_0\log p}{n}\right)
\end{eqnarray}
and
\begin{eqnarray}
\label{yu8}
n^{-1}\sum_{t=1}^{n}(\mathbf{x}_t^{\top}(\hat{\bm{\beta}}(\hat{J}_{\hat{k}_n})-\bm{\beta}^{*}))^{2}= O_{p}
\left(\frac{k_0\log p}{n}\right).
\end{eqnarray}
Although \eqref{yu5}--\eqref{yu8} suggest that Lasso and OGA+HDAIC
share the same error rate in the case of $k_0 \ll n$,
they are obtained under somewhat different assumptions.
Note first that unlike \eqref{yu5} and \eqref{yu6},
\eqref{yu7} and \eqref{yu8} do not require that $\{\mathbf{x}_t\}$ and $\{\varepsilon_t\}$
are independent, and hence are applicable to ARX models.
Moreover, \eqref{yu5} and \eqref{yu6} are established under
\begin{align}
\label{yu9}
\displaystyle{\mathrm{ess \,sup}_{\theta \in [-\pi, \pi]}}\lambda_{\max}(f_{\mathbf{x}}(\theta)) < \bar{S},
\end{align}
and
\begin{eqnarray}
\label{yu101}
\displaystyle{\mathrm{ess \,inf}_{\theta \in [-\pi, \pi]}}\lambda_{\min}(f_{\mathbf{x}}(\theta)) > \underline{s},
\end{eqnarray}
where $0<\underline{s}\leq \bar{S}<\infty$
and
$f_{\mathbf{x}}(\theta)=[1/(2\pi)]\sum_{l=-\infty}^{\infty}\bm{\Gamma}_{\mathbf{x}}(l)\exp(-il\theta)$
with $\bm{\Gamma}_{\mathbf{x}}(l)=\mathrm{E}(\mathbf{x}_{t}\mathbf{x}_{t+l}^{\top})$.
Assumption \eqref{yu101} is comparable to \eqref{ts10} (which
assumes that $\lambda_{\min}(\bm{\Gamma})$ is bounded away
from zero and is needed for
proving \eqref{yu7} and \eqref{yu8}), but is more stringent than the latter
because
\begin{align*}
\begin{split}
& \lambda_{\min}(\bm{\Gamma})=\lambda_{\min}(\bm{\Gamma}_{\mathbf{x}}(0))=
\lambda_{\min}\left(\int_{-\pi}^{\pi}f_{\mathbf{x}}(\theta)d \theta\right) \\
&\geq
2\pi\{\displaystyle{\mathrm{ess \,inf}_{\theta \in [-\pi, \pi]}}\lambda_{\min}(f_{\mathbf{x}}(\theta))\}.
\end{split}
\end{align*}
Maximum eigenvalue assumptions like \eqref{yu9} are not required for
\eqref{yu7} and \eqref{yu8}. This type of assumption can be easily
violated when the components of $\mathbf{x}_t$ are highly
correlated, as illustrated by an ARX example in Section S3
of the supplementary document, in which $\lambda_{\max}(\bm{\Gamma})
\to \infty$ as $p \to \infty$ and hence $\mathrm{ess \,sup}_{\theta
\in [-\pi, \pi]} \lambda_{\max}(f_{\mathbf{x}}(\theta))\geq
[1/(2\pi)]\lambda_{\max}(\bm{\Gamma})$. On the other hand, while
\eqref{yu7} and \eqref{yu8} are obtained under the beta-min
condition given in \eqref{sep6}, \eqref{yu5} and \eqref{yu6} do not
assume any beta-min condition. Wu and Wu (2016) also investigate the
performance of Lasso under \eqref{yu1} with $k_0 \ll n$ and
$\{x_{ti}\}$ being nonrandom and obeying the restricted eigenvalue
condition defined in (4.2) of their paper. They allow
$\{\epsilon_t\}$ to be a stationary process following some general
moment and dependence conditions. The error rates that they derive
for Lasso, however, are usually larger than those in
\eqref{yu5}--\eqref{yu8}.

In fact, it can be argued that all error bounds obtained in Theorem
\ref{thm3.1} are rate optimal. To see this, let $\hat{J}(m), 1\leq m \leq K_n $,
be a sequence of nested models chosen
from $p$ candidate variables in a data-driven fashion,
where
$\sharp(\hat{J}(m))=m$.
The CMSPE of model $\hat{J}(m)$ is
$\mathrm{E}_n(y(\mathbf{x})-\hat{y}_{\hat{J}(m)}(\mathbf{x}))^2=
\mathrm{E}_n(y(\mathbf{x})-y_{\hat{J}(m)}(\mathbf{x}))^2+\mathrm{E}_n(\hat{y}_{\hat{J}(m)}(\mathbf{x})-y_{\hat{J}(m)}(\mathbf{x}))^2
$, where $\hat{y}_{J}(\mathbf{x})=\mathbf{x}^{\top}(J)
\hat{\bm{\beta}}(J)$. It is not difficult to show that
the squared bias terms obey
\begin{eqnarray}
\label{rev21a}
\mathrm{E}_n(y(\mathbf{x})-y_{\hat{J}(m)}(\mathbf{x}))^2 \geq
\mathrm{E}(y(\mathbf{x})-y_{J_m^{*}}(\mathbf{x}))^2,
\end{eqnarray}
where $y_{J_m^{*}}(\mathbf{x})$, satisfying
$\sharp(J_m^{*})=m$ and
\begin{align}
\label{may19}
\mathrm{E}(y(\mathbf{x})-y_{J^{*}_{m}}(\mathbf{x}))^{2} =
\min_{\sharp(J)=m}\mathrm{E}(y(\mathbf{x})-y_{J}(\mathbf{x}))^{2},
\end{align}
is called the best $m$-term approximation of
$y(\mathbf{x})$.
In addition, an argument
similar to that used to prove \eqref{ts28} implies
that the variance terms
satisfy
\begin{eqnarray}
\label{rev18b}
\max_{1\leq m \leq K_n}
\frac{n\mathrm{E}_n(\hat{y}_{\hat{J}(m)}(\mathbf{x})-y_{\hat{J}(m)}(\mathbf{x}))^2}{m \log p}=O_{p}(1).
\end{eqnarray}
In view of \eqref{rev21a} and \eqref{rev18b},
the best possible rate that can be achieved by a forward inclusion method
accompanied by a stopping criterion
is the same as that of
\begin{align}
\label{rev18c}
\begin{split}
\bar{L}_{n}(m_n^*)\equiv \min_{1\leq m \leq K_n}\bar{L}_n(m)=\min_{1\leq m \leq K_n}
\{\mathrm{E}(y(\mathbf{x})-y_{J_m^{*}}(\mathbf{x}))^2+m\log p/n\}.
\end{split}
\end{align}
According to
Lemma \ref{l3}, \eqref{rev1}, and $\mathrm{E}(y(\mathbf{x})-y_{J_m^{*}}(\mathbf{x}))^2=0$ if $m\geq k_0$,
the convergence rate of $\bar{L}_{n}(m_n^*)$
under \eqref{ts5}, \eqref{intro3}, or \eqref{aug13}
is $(\log p/n)^{1-1/2\gamma}$, $\log n \log p/n,$
or $k_0\log p/n$, which coincides with that of
\eqref{3-3}, \eqref{3-2}, or \eqref{3-1}, respectively.
We therefore conclude that
the bounds obtained in Theorem \ref{thm3.1} are rate optimal.
In this connection, we also note that when \eqref{aug1} is a stationary AR($p$) model with $p \gg n$,
the set of candidate models are usually given by
AR(1)$, \ldots,$ AR($K_n$), with $K_n$ approaching
$\infty$ at a rate slower than $n$.
Unlike $\hat{J}(m), 1\leq m \leq K_n$,
the candidate set in this case is not determined by any data-driven methods,
and hence
the corresponding variance terms can get rid of the variance inflation factor $\log p$ (see \eqref{rev18b}),
which is introduced by data-dependent selection of the candidate set from all $p$ variables.
As a result, the optimal rate that can be attained by an order selection criterion
is equivalent to that of
\begin{align}
\label{rev21b}
\begin{split}
\min_{1\leq m \leq K_n}
\{\mathrm{E}(y(\mathbf{x})-y_{J_m^{*}}(\mathbf{x}))^2+m/n\};
\end{split}
\end{align}
see Shibata (1980) for more details.
Under
\eqref{intro1}, \eqref{intro2}, or
\eqref{aug13} with $N_n=\{1, \ldots, k_0\}$,
the convergence rate of
\eqref{rev21b} is
$(1/n)^{1-1/2\gamma}$, $\log n/n$, or
$k_0/n$, which differs by a factor of
$(\log p)^{1-1/2\gamma}$ from that
of $\bar{L}_{n}(m_n^*)$
under
\eqref{ts5}, \eqref{intro3}, or \eqref{aug13}, respectively.

We would also like to point out the differences between the current paper
and the paper by Ing and Lai (2011), which investigates the performance of
OGA under \eqref{aug1} with
$(\mathbf{x}_t, \varepsilon_t)$ being i.i.d. and obeying
sub-Gaussian or subexponential distributions.
Note first that
Theorem 1 of Ing and Lai (2011)
can be understood as a special case of Theorem \ref{t1}
when $\gamma=1$ and observations are independent over time.
However, since the former theorem only focuses on the case of $\gamma=1$,
its proof does not involve the approximation errors of the population OGA
under general sparsity conditions such as those given in Lemma \ref{l1} in the Appendix.
Moreover, when $\gamma=1$ is known,
the optimal rate, $(\log p/n)^{1/2}$,
can be achieved by choosing $m=(n/\log p)^{1/2}$,
without recourse to any data-driven method to help determine the number of iterations.
Alternatively, Theorem \ref{thm3.1} encompasses a much wider class of sparsity conditions,
and demonstrates that
HDAIC can automatically
choose a suitable $m$,
leading to the optimal balance between the squared bias term and the variance term,
without knowing the degree of sparseness.
Indeed, Theorem 4 of Ing and Lai (2011) has suggested
using a high-dimensional information criterion (whose penalty is
heavier than that of HDAIC) to decide the number of OGA iterations when the regression coefficients satisfy the
strong sparsity condition, \eqref{aug13},
and a beta-min condition.
Theorem 5 of Ing and Lai (2011) further introduces a backward elimination method
based on the aforementioned information criterion
to remove possible redundant variables surviving the first two (variable) screening stages,
and shows that the resultant set of variables is equivalent to $N_n$ with probability tending to 1.
Although the approaches adopted in both papers can be considered similar to a certain extent,
their goals are entirely different.
In particular, whereas Ing and Lai (2011) aim to establish selection consistency under
the strong sparsity condition, this paper focuses on
prediction efficiency under much more general sparsity conditions,
which include the strong sparsity one as a special case.
From a technical point of view, the main differences between the two papers are:
(i) serial correlation is not allowed in Ing and Lai (2011); and (ii)
the squared bias term in Theorems 4 (or Theorem 5) of Ing and Lai (2011)
completely vanishes along the OGA path in the sense that
\begin{align*}
\begin{split}
P\big(\min_{1\leq m \leq K_n}
\mathrm{E}_n(y(\mathbf{x})-y_{\hat{J}_m}(\mathbf{x}))^2=0\big) \to 1, \quad \mbox{as} \,\, n \to \infty,
\end{split}
\end{align*}
which is ensured by
the sure screening property of OGA under the strong sparsity condition (see Theorem 3 of Ing and Lai (2011)),
but
the squared bias term in Theorem \ref{thm3.1} decays
at a variety of unknown rates and can never be zero along the OGA path,
making it much harder to pursue
the bias-variance tradeoff
along this data-driven path.


We close this section by mentioning that
while condition \eqref{hdic1} on $s_a$ involves unknown parameters,
we have introduced a data-driven method for determining $s_a$ in
Section S3 of the supplementary document,
which is of practical relevance.

\subsection{Proof of Theorem \ref{thm3.1}}
\label{sec3.2}
We only prove \eqref{3-3}.
The proof of
\eqref{3-2} is similar to that of \eqref{3-3}, and hence is omitted.
The proof of \eqref{3-1} is slightly different,
and is deferred to the supplementary material
because of space constraints.
In the rest of the proof,
a weaker restriction on the penalty term,
\begin{eqnarray}
\label{hdic2}
s_a>\bar{V}^{*}\equiv\frac{2\bar{B}c_1^{*^2}}{\sigma^{2}},
\end{eqnarray}
is used instead of \eqref{hdic1}, although the latter one
is required in the proof of \eqref{3-1}.

By making use of \eqref{ts3},
\eqref{ubc} (which is ensured by \eqref{ts8}),
and \eqref{ts10}, we show in Section \ref{app1} in the Appendix that
for any $1\leq m \leq K_n$,
\begin{align}
\label{31}
\begin{split}
& -C_{M, \gamma, \lambda_1}R_{1,p} \{\mathrm{E}_n(\varepsilon^2(\hat{J}_m))\}^{(2\gamma-2)/(2\gamma-1)}
\leq n^{-1}\sum_{t=1}^{n}\varepsilon^2_{t}(\hat{J}_m)- \mathrm{E}_n(\varepsilon^2(\hat{J}_m))\\
&\leq  C_{M, \gamma, \lambda_1}R_{1,p} \{\mathrm{E}_n(\varepsilon^2(\hat{J}_m))\}^{(2\gamma-2)/(2\gamma-1)},
\end{split}
\end{align}
where
$C_{M, \gamma, \lambda_1}=(M+1)^2C^2_\gamma \lambda_1^{-(2\gamma-2)/(2\gamma-1)}$,
with $M$ defined in \eqref{ts8},
$R_{1,p}=\max_{1\leq i,l\leq p}|n^{-1}\sum_{t=1}^{n}z_{ti}z_{tl}-\rho_{il}|$,
$\varepsilon_{t}(J)=y_t-\varepsilon_t-\bm{\beta}^{*^{\top}}(J)\mathbf{z}_{t}(J)$, and
$\varepsilon (J)=y(\mathbf{x})-y_{J}(\mathbf{x})=y(\mathbf{x})-\bm{\beta}^{*^{\top}}(J)\mathbf{z}(J)$.
In addition, it is shown in Section S2 of the supplementary material that
\begin{align}
\label{32}
\begin{split}
& |n^{-1}\sum_{t=1}^{n}\varepsilon_t\varepsilon_{t}(\hat{J}_m)| \leq C^{1/2}_{M, \gamma, \lambda_1}
R_{2,p}\{\mathrm{E}_n(\varepsilon^2(\hat{J}_m))\}^{(\gamma-1)/(2\gamma-1)},
\end{split}
\end{align}
\begin{align}
\label{33}
\begin{split}
\max_{1\leq m \leq K_n}\frac{\|n^{-1}\sum_{t=1}^{n}\mathbf{z}_t(\hat{J}_m)
\varepsilon_{t}(\hat{J}_m)\|^2_{\hat{\bm{\Gamma}}^{-1}(\hat{J}_m)}}
{m\{\mathrm{E}_n(\varepsilon^2(\hat{J}_m))\}^{(2\gamma-2)/(2\gamma-1)}} \leq C_{M, \gamma, \lambda_1}\|\hat{\bm{\Gamma}}^{-1}(\hat{J}_{K_n})\|
R_{1,p}^2,
\end{split}
\end{align}
and
\begin{align}
\label{331}
\begin{split}
\max_{1\leq m \leq K_n}\frac{\|n^{-1}\sum_{t=1}^{n}\mathbf{z}_t(\hat{J}_m)
\varepsilon_{t}\|^2_{\hat{\bm{\Gamma}}^{-1}(\hat{J}_m)}}
{m} \leq \|\hat{\bm{\Gamma}}^{-1}(\hat{J}_{K_n})\| R_{2,p}^2,
\end{split}
\end{align}
where $R_{2,p}= \max_{1\leq i \leq p}|n^{-1}\sum_{t=1}^{n}z_{ti}\varepsilon_t|$
and
$\|\bm{\nu}\|^{2}_{\bm{A}}=\bm{\nu}^{\top}\bm{A}\bm{\nu}$ for vector $\bm{\nu}$
and non-negative definite matrix $\bm{A}$.

Let $m^*_n=\min\{(n/\log p)^{1/2\gamma}, K_n\}$ and
\begin{align}
\label{34}
\begin{split}
\tilde{k}_n=\min\{k: 1 \leq k \leq K_n, \mathrm{E}_n(\varepsilon^2(\hat{J}_k))\leq Gm^{*^{-2\gamma+1}}_n \} \,\,(\min \emptyset =K_n),
\end{split}
\end{align}
in which
$G\gg C_2$ and $C2$ is defined in \eqref{ts26}.
Using \eqref{31}--\eqref{331}, we next show that
\begin{eqnarray}
\label{35}
\lim_{n \to \infty}P(\hat{k}_n \leq \tilde{k}_n-1)=0.
\end{eqnarray}

Since \eqref{ts26} implies
$
\mathrm{E}_{n}(\varepsilon^{2}(\hat{J}_{m_n^{*}}))\leq C_2
m_{n}^{*^{-2\gamma+1}}\leq Gm^{*^{-2\gamma+1}}_n$ on $A_n(K_n)$,
it follows that
$m^{*}_n \geq \tilde{k}_n$ on $A_n(K_n)$.
By \eqref{ts-1}, one obtains
\begin{align}
\label{37}
\begin{split}
&P(\hat{k}_{n} \leq \tilde{k}_n-1) \leq P(\hat{k}_{n} \leq \tilde{k}_n-1, A_n(K_n))+P(A^{c}_n(K_n))\\
&\leq P\left(\min_{1\leq k \leq \tilde{k}_n-1}Q_n(k) \leq s_am_n^{*}(n^{-1}\sum_{t=1}^{n}y_{t}^{2})\log p/n, A_n(K_n)\right)+o(1),
\end{split}
\end{align}
where
\begin{align*}
\begin{split}
& Q_n(k)=n^{-1}\sum_{t=1}^{n}\varepsilon_{t}^{2}(\hat{J}_k)+2n^{-1}
\sum_{t=1}^{n}\varepsilon_{t}(\hat{J}_k)\varepsilon_t-2n^{-1}
\sum_{t=1}^{n}\varepsilon_{t}(\hat{J}_{m_n^{*}})\varepsilon_t \\
&-n^{-1}\sum_{t=1}^{n}\varepsilon_{t}^{2}(\hat{J}_{m_n^{*}})-
\|n^{-1}\sum_{t=1}^{n}\mathbf{z}_{t}(\hat{J}_{k})(\varepsilon_t+\varepsilon_{t}(\hat{J}_k))\|^2_{\hat{\bm{\Gamma}}^{-1}(\hat{J}_k)}.
\end{split}
\end{align*}

By \eqref{ts1.1}, \eqref{ts2.1}, and \eqref{332},
\begin{eqnarray}
\label{38}
\lim_{n \to \infty} P(W_n)=1,
\end{eqnarray}
where
\begin{align*}
\begin{split}
&W_n= \{R_{1, p}\leq c_2^* (\log p)^{1/2}/n^{1/2}\}\bigcap\{R_{2, p}\leq c_1^* (\log p)^{1/2}/n^{1/2}\} \\
&\bigcap \{\|\hat{\bm{\Gamma}}^{-1}(\hat{J}_{K_n})\| \leq \bar{B}\}.
\end{split}
\end{align*}
Moreover, \eqref{31}--\eqref{331}, \eqref{332}, and \eqref{ts26} imply that for $1
\leq k \leq \tilde{k}_n-1$ and all large $n$,
\begin{align}
\label{39}
\begin{split}
&n^{-1}\sum_{t=1}^{n}\varepsilon^2_{t}(\hat{J}_k) \geq \mathrm{E}_n(\varepsilon^2(\hat{J}_k)) \\
&\times \left\{1- \frac{C_{M, \gamma, \lambda_1}c_2^{*}}{G^{1/(2\gamma-1)}}
\big(I_{\{\gamma=1\}}+
\big(\frac{\log p}{n}\big)^{(\gamma-1)/2\gamma}I_{\{\gamma>1\}}\big)\right\}
\,\,\, \mbox{on}\,\,W_n,
\end{split}
\end{align}
\begin{align}
\label{40}
\begin{split}
n^{-1}|\sum_{t=1}^{n}\varepsilon_{t}(\hat{J}_k)\varepsilon_t| \leq \mathrm{E}_n(\varepsilon^2(\hat{J}_k))
\frac{C^{1/2}_{M, \gamma, \lambda_1}c_1^{*}}{G^{\gamma/(2\gamma-1)}} \,\,\, \mbox{on}\,\,W_n,
\end{split}
\end{align}
\begin{align}
\label{41}
\begin{split}
n^{-1}|\sum_{t=1}^{n}\varepsilon_{t}(\hat{J}_{m_n^{*}})\varepsilon_t| \leq \mathrm{E}_n(\varepsilon^2(\hat{J}_k))
\frac{C^{1/2}_{M, \gamma, \lambda_1}c_1^{*}}{G^{\gamma/(2\gamma-1)}} \,\,\, \mbox{on}\,\,W_n\bigcap A_n(K_n),
\end{split}
\end{align}
\begin{align}
\label{Jun1}
\begin{split}
& n^{-1}\sum_{t=1}^{n}\varepsilon_{t}^2(\hat{J}_{m_n^{*}}) \leq \mathrm{E}_n(\varepsilon^2(\hat{J}_k)) \\
&\times \left\{\frac{C_2}{G}+
\frac{C_{M, \gamma, \lambda_1}c_2^{*}}{G^{1/(2\gamma-1)}}
\big(I_{\{\gamma=1\}}+
\big(\frac{\log p}{n}\big)^{(\gamma-1)/2\gamma}I_{\{\gamma>1\}}\big)\right\}
\,\,\, \mbox{on}\,\,W_n\bigcap A_n(K_n),
\end{split}
\end{align}
\begin{align}
\label{42}
\begin{split}
&\|n^{-1}\sum_{t=1}^{n}\mathbf{z}_t(\hat{J}_k) \varepsilon_{t}(\hat{J}_k)
\|^2_{\hat{\bm{\Gamma}}^{-1}(\hat{J}_k)}
\leq \mathrm{E}_n(\varepsilon^2(\hat{J}_k))\\
&\times \frac{C_{M, \gamma, \lambda_1}\bar{B}
c_2^{*^2}\bar{\delta}}{G^{1/(2\gamma-1)}}\big(I_{\{\gamma=1\}}+
\big(\frac{\log p}{n}\big)^{(\gamma-1)/2\gamma}I_{\{\gamma>1\}}\big)
 \,\,\, \mbox{on}\,\,W_n\bigcap A_n(K_n),
\end{split}
\end{align}
and
\begin{align}
\label{43}
\begin{split}
&\|n^{-1}\sum_{t=1}^{n}\mathbf{z}_t(\hat{J}_k)
\varepsilon_{t}\|^2_{\hat{\bm{\Gamma}}^{-1}(\hat{J}_k)}
\leq \mathrm{E}_n(\varepsilon^2(\hat{J}_k))
\frac{\bar{B} c_1^{*^2}}{G} \,\,\, \mbox{on}\,\,W_n\bigcap A_n(K_n).
\end{split}
\end{align}
By \eqref{39}--\eqref{43}, it follows that
for large enough $G$ in \eqref{34}, there exists $0<\iota<1/2$ such that for all large $n$,
\begin{align}
\label{44}
\begin{split}
& \min_{1 \leq k \leq \tilde{k}_{n}-1}Q_{n}(k) \geq \min_{1 \leq k \leq \tilde{k}_{n}-1}
\mathrm{E}_n(\varepsilon^2(\hat{J}_k))(1-\iota) \\
&\geq Gm^{*^{-2\gamma+1}}_{n}(1-\iota)\,\,\, \mbox{on}\,\,W_n\bigcap A_n(K_n).
\end{split}
\end{align}
In addition, \eqref{ts1.1}, \eqref{ts2.1}, (A3), and $\log p/n \leq m^{*^{-2\gamma+1}}_n$
ensure that there exists $\bar{M}_2>0$ such that
\begin{eqnarray}
\label{45}
\lim_{n \to \infty}
P\left(\frac{s_a m_n^{*}n^{-1}\sum_{t=1}^{n}y_t^2}{n}\log p \leq \bar{M}_2m^{*^{-2\gamma+1}}_{n}\right)=1.
\end{eqnarray}
By \eqref{ts-1}, \eqref{38}, \eqref{45}, and selecting $G$ in \eqref{44}
larger than $2\bar{M}_2$, we obtain the desired conclusion \eqref{35}.

Using \eqref{31}--\eqref{331} again,
it is shown in Section S2 of the supplementary material that
\begin{eqnarray}
\label{36}
\lim_{n \to \infty}P(\hat{k}_n \geq Vm^{*}_n)=0, \gamma>1,
\end{eqnarray}
where $V$ is a sufficiently large constant to be specified in the proof of \eqref{36}.
With the help of \eqref{35} and \eqref{36}, the desired conclusion follows if one can show that
for $\gamma>1$,
\begin{align}
\label{55}
\begin{split}
\mathrm{E}_n\{y(\mathbf{x})-\hat{y}_{\hat{k}_n}(\mathbf{x})\}^2I_{\{\tilde{k}_n \leq \hat{k}_n < Vm_{n}^{*}\}}=
O_{p}(m_n^{*^{-2\gamma+1}}),
\end{split}
\end{align}
and for $\gamma=1$,
\begin{align}
\label{555}
\begin{split}
\mathrm{E}_n\{y(\mathbf{x})-\hat{y}_{\hat{k}_n}(\mathbf{x})\}^2I_{\{\tilde{k}_n \leq \hat{k}_n \leq K_n\}}=
O_{p}\big((\log p/n)^{1/2}\big).
\end{split}
\end{align}
To show \eqref{55}, note first that
\begin{align}
\label{56}
\begin{split}
&\mathrm{E}_{n}(y(\mathbf{x})-\hat{y}_{\hat{k}_n}(\mathbf{x}))^2I_{\{\tilde{k}_n\leq \hat{k}_n \leq Vm_n^{*}\}} \\
&\leq
\mathrm{E}_{n}\varepsilon^2(\hat{J}_{\tilde{k}_n})+
\|\mathbf{L}(\hat{J}_{\hat{k}_n})\|^2\|\hat{\bm{\Gamma}}^{-1}(\hat{J}_{\hat{k}_n})\|I_{\{\tilde{k}_n\leq \hat{k}_n \leq Vm_n^{*}\}}\\
&+ \|\mathbf{L}(\hat{J}_{\hat{k}_n})\|^2
\|\hat{\bm{\Gamma}}^{-1}(\hat{J}_{\hat{k}_n})\|^2
\|\hat{\bm{\Gamma}}(\hat{J}_{\hat{k}_n})-\bm{\Gamma}(\hat{J}_{\hat{k}_n})\|I_{\{\tilde{k}_n\leq \hat{k}_n \leq Vm_n^{*}\}},
\end{split}
\end{align}
where
$\mathbf{L}(J)=n^{-1}\sum_{t=1}^{n}\mathbf{z}_{t}(J)(\varepsilon_t+\varepsilon_t(J))$.
By (A3), \eqref{ts1.1}, \eqref{ts2.1}, \eqref{ts8}, and straightforward algebraic manipulations,
it holds that
\begin{align}
\label{57}
\begin{split}
& \|\mathbf{L}(\hat{J}_{\hat{k}_n})\|^{2}I_{\{\tilde{k}_n\leq \hat{k}_n \leq Vm_n^{*}\}}\\
&\leq
2Vm_n^{*}\max_{1\leq i \leq p}(n^{-1}\sum_{t=1}^{n}\varepsilon_{t}z_{ti})^{2} +
2Vm_n^{*}\max_{1\leq i, j \leq p}
(n^{-1}\sum_{t=1}^{n}z_{ti}z_{tj}-\rho_{ij})^{2} \\
& \times
(\sum_{j=1}^{p}|\beta^{*}_{j}|)^{2}
(1+\max_{1 \leq \sharp(J) \leq K_{n}, 1 \leq l \leq p}
\|\mbox{\boldmath$\Gamma$}^{-1}(J)\mathbf{g}_{l}(J)\|_{1})^{2}= O_{p}\left(\frac{m_n^{*}\log p}{n}\right)\\
&=  O_{p}\left(m_n^{*^{-2\gamma+1}} \right).
\end{split}
\end{align}
Moreover, we have
\begin{align}
\label{58}
\begin{split}
\mathrm{E}_n\varepsilon^2(\hat{J}_{\tilde{k}_n}) \leq
\mathrm{E}_n\varepsilon^2(\hat{J}_{m^{*}_n}) \leq
 C_2m_n^{*^{-2\gamma+1}} \,\,\mbox{on} \,\, A(K_n),
\end{split}
\end{align}
and
\begin{align}
\label{59}
\begin{split}
&\|\hat{\bm{\Gamma}}(\hat{J}_{\hat{k}_n})-\bm{\Gamma}(\hat{J}_{\hat{k}_n})\|I_{\{\tilde{k}_n\leq \hat{k}_n \leq Vm_n^{*}\}}
\leq K_n\max_{1\leq i, j \leq p}|n^{-1}\sum_{t=1}^{n}z_{ti}z_{tj}-\rho_{ij}|\\
&=O_{p}(1),
\end{split}
\end{align}
where the equality is ensured by \eqref{ts2.1} and \eqref{kn}.
Consequently, \eqref{55} follows from \eqref{56}--\eqref{59}, \eqref{ts-1},
and \eqref{332}.
The proof of \eqref{555} is similar to that of \eqref{55}.
The details are omitted.
\hfill{$\Box$}

\section{Conclusions}
\label{sec4}
\indent
This paper has addressed the important problem of
selecting
high-dimensional linear regression models with
dependent observations
when knowledge is lacking about the
degree of sparseness of the true model.
When the true model is known to be an AR model
or a regression model whose predictor variables have been ranked a priori based on their importance,
this type of problem has been tackled in the past; see, e.g., Ing (2007), Yang (2007),
Zhang and Yang (2015), and Ding et al. (2018).
These authors have proposed various ways to combine
the strengths of AIC and BIC and shown that their methods
achieve the optimal rate
without knowing whether \eqref{intro1}, \eqref{intro2}, or
\eqref{aug13}, with $N_n=\{1, \ldots, k_0\}$,
is true. Their approaches, however, are not applicable to situations where
the predictor variables have no natural ordering
or their importance ranks are unknown.
To alleviate this difficulty,
we first use OGA to rank predictor variables,
and then choose along the OGA path the model
that has the smallest HDAIC value.
Our approach is not only computationally feasible, but also rate optimal
without the need for knowing how sparse the underlying time series model is.

Compared to a similar attempt made in Negahban et al. (2012),
in which Lasso is used instead of OGA+HDAIC, the novelty of this paper is threefold:
first, the validity of OGA+HDAIC is established
not only for independent data, but also for time series data; second,
the advantage of OGA+HDAIC is obtained
in the important special case \eqref{intro3},
which is seldom discussed in the high-dimensional literature;
third, in another
important special case \eqref{ts5},
it is shown that
OGA+HDAIC can have a faster convergence rate than Lasso.
Finally, we note that OGA is exclusive for linear models.
The counterpart of OGA in nonlinear models is the Chebyshev greedy algorithm (CGA) (Temlyakov, 2015).
Investigating the performance of CGA+HDAIC
in high-dimensional nonlinear time series models
would be an interesting topic for future research.




\setcounter{equation}{0}
\setcounter{section}{0}
\setcounter{table}{0}
\setcounter{figure}{0}
\def\theequation{A\arabic{section}.\arabic{equation}}
\def\thetable{A\arabic{section}.\arabic{table}}
\def\thefigure{A\arabic{section}.\arabic{figure}}
\def\thesection{A\arabic{section}}
\fontsize{12}{14pt plus.8pt minus .6pt}\selectfont
\begin{flushleft}
{\Large\bf Appendix}
\end{flushleft}
\section{Rates of Convergence of the Population OGA}
\label{sec2.2}
In this section,
we consider the population counterpart of OGA,
whose convergence rate plays a crucial role in the analysis of
the first term on the right-hand side of \eqref{aug14}.
Let $0<\xi\leq 1$ be given. The algorithm initializes $J_{\xi, 0}=\emptyset$.
For $m \geq 1$, $J_{\xi, m}$ is recursively updated by
\begin{eqnarray*}
J_{\xi, m}=J_{\xi, m-1} \bigcup \{j_{\xi, m}\},
\end{eqnarray*}
where
$j_{\xi, m}$ is any element $l$ in ${\cal P}$ satisfying
\begin{align*}
\begin{split}
|\mathrm{E}(u_{m-1}z_{l})|\geq
 \xi \max_{1\leq j \leq p}|\mathrm{E}(u_{m-1}z_{j})|,
\end{split}
\end{align*}
with $u_{0}=y(\mathbf{x})$ and
$u_{m}=y(\mathbf{x})-y_{J_{\xi, m}}(\mathbf{x})$ if $ m \geq 1$.
Because the algorithm
is implemented based on
the `population' correlations
of $\mathbf{x}$,
it is referred to as the population OGA when $\xi=1$,
and the population weak OGA when $0<\xi<1$.
The following lemma provides a rate of convergence
of the $\mathrm{E}(u^2_{m})$ under (A3) or (A4).

\begin{lam}
\label{l1}
Assume \eqref{ts3} and
\eqref{ts10}.
Then, there exists $G_{1}>0$ such that
\begin{align}
\label{ts11}
\mathrm{E}(u^2_{m})=\mathrm{E}(y(\mathbf{x})-y_{J_{\xi, m}}(\mathbf{x}))^{2} \leq G_{1} m^{-2 \gamma +1}.
\end{align}
Moreover, if \eqref{ts6} holds instead of \eqref{ts3},
then there exist $G_{2}, G_{3}>0$ such that
\begin{align}
\label{ts12}
\mathrm{E}(u^2_{m})=\mathrm{E}(y(\mathbf{x})-y_{J_{\xi, m}}(\mathbf{x}))^{2} \leq G_{2} {\rm exp}(-G_{3} m).
\end{align}
\end{lam}

\noindent
{\bf Proof.}
Straightforward calculations yield
\begin{align}
\label{ts13}
\begin{split}
& \mathrm{E}(u^2_{m})= \mathrm{E}\big[(y(\mathbf{x})-y_{J_{\xi, m}}(\mathbf{x}))\sum_{j=1}^{p}\beta_{j}^*z_{j}\big] \\
&\leq \max_{1 \leq j \leq p}|\mu_{J_{\xi, m}, j}|\sum_{j=1, j \notin J_{\xi, m}}^{p}|\beta_{j}^*|,
\end{split}
\end{align}
recalling $\mu_{J, i}=\mathrm{E}[(y(\mathbf{x})-y_{J}(\mathbf{x}))z_{i}]$.
In addition, \eqref{ts10} implies
\begin{align}
\label{ts14}
\mathrm{E}(u^2_{m}) \geq \lambda_{1}\sum_{j=1, j \notin J_{\xi, m}}^{p}
\beta_{j}^{*^2}.
\end{align}
By \eqref{ts13}, \eqref{ts14} and \eqref{ts3}, it follows that
\begin{align}
\label{ts0}
\begin{split}
& \mathrm{E}(u^2_{m}) \leq
C_{\gamma}\max_{1 \leq j \leq p}|\mu_{J_{\xi, m}, j}|
(\sum_{j=1, j \notin J_{\xi, m}}^{p}\beta_{j}^{*^2})^{(\gamma-1)/(2\gamma-1)} \\
&\leq C_{\gamma} \lambda_{1}^{-(\gamma-1)/(2\gamma-1)}
\max_{1 \leq j \leq p}|\mu_{J_{\xi, m}, j}|
[\mathrm{E}(u^2_{m})]^{(\gamma-1)/(2\gamma-1)},
\end{split}
\end{align}
and hence
\begin{align}
\label{ts15}
& [\mathrm{E}(u^2_{m})]^{\gamma/(2\gamma-1)}
\leq C_{\gamma} \lambda_{1}^{-(\gamma-1)/(2\gamma-1)}
\max_{1 \leq j \leq p}|\mu_{J_{\xi, m}, j}|.
\end{align}
In view of \eqref{ts15}, one has
\begin{align}
\label{ts16}
\begin{split}
& \mathrm{E}(u^{2}_{m+1}) \leq \mathrm{E}(u_{m}-\mu_{J_{\xi, m}, j_{\xi, m+1}}z_{j_{\xi, m+1}})^{2} \\
& \leq \mathrm{E}(u^{2}_{m})-\xi^{2}\max_{1 \leq j \leq p}\mu^{2}_{J_{\xi, m}, j} \\
& \leq \mathrm{E}(u^{2}_{m})-\xi^{2} \lambda^{2(\gamma-1)/(2\gamma-1)}_{1}C^{-2}_{\gamma}
[\mathrm{E}(u^{2}_{m})]^{2\gamma/(2\gamma-1)} \\
&=
 \mathrm{E}(u^{2}_{m})\{1-\xi^{2} \lambda^{2(\gamma-1)/(2\gamma-1)}_{1}C^{-2}_{\gamma}
[\mathrm{E}(u^{2}_{m})]^{1/(2\gamma-1)}\}.
\end{split}
\end{align}
The desired conclusion \eqref{ts11} follows from \eqref{ts16} and Lemma 1 of Gao et al. (2013).

To show \eqref{ts12}, note first that \eqref{ts6}, \eqref{ts13} and \eqref{ts14} yield
\begin{align*}
\mathrm{E}(u_{m}^{2})^{1/2} \leq \lambda^{-1/2}_{1}M_1\max_{1 \leq j \leq p}|\mu_{J_{\xi, m}, j}|.
\end{align*}
This and an argument similar to that used in \eqref{ts16} imply
\begin{align}
\label{ts17}
\begin{split}
& \mathrm{E}(u^{2}_{m+1})
\leq \mathrm{E}(u^{2}_{m})-\xi^{2} \lambda_{1}M^{-2}_1 \mathrm{E}(u^{2}_{m}) \\
&=
 \mathrm{E}(u^{2}_{m})\{1-\xi^{2} \lambda_{1}M^{-2}_1\}.
\end{split}
\end{align}
Since $M_1>1$, $0<\lambda_1 \leq 1$,
and $0<\xi\leq 1$, \eqref{ts17} leads directly to \eqref{ts12}.
\hfill$\Box$

\vspace{0.1cm}
Lemma \ref{l2} shows that \eqref{ts5} is a special case of \eqref{ts3}.
Using Lemmas \ref{l1} and \ref{l2},
Lemma \ref{l3} demonstrates that the rate $m^{-2\gamma+1}$
obtained in \eqref{ts11} cannot be improved under \eqref{ts5}.
More specifically, recall
the best $m$-term approximation, $y_{J^{*}_{m}}(\mathbf{x})$, of $y(\mathbf{x})$ (see \eqref{may19}).
Lemma \ref{l3} asserts that
when \eqref{ts5} and \eqref{ts10} hold true,
the approximation errors of
$y_{J_{\xi, m}}(\mathbf{x})$ and $y_{J^{*}_{m}}(\mathbf{x})$
only differ by a positive constant.

\begin{lam}
\label{l2}
Suppose that \eqref{ts5} is true for some $\gamma>1$. Then \eqref{ts3} holds for the same $\gamma$.
\end{lam}

\noindent
{\bf Proof.} See Section S1 of the supplementary document.

\begin{lam}
\label{l3}
Suppose that \eqref{ts5} holds for some $\gamma>1$ and \eqref{ts10} is true. Then,
for all $1\leq m \leq (1-\epsilon)p$, where $\epsilon$
is an arbitrarily small positive constant, there exist $D_{1}, D_{2}$ and $D_{3}$ such that
\begin{align}
\label{ts18}
\mathrm{E}(y(\mathbf{x})-y_{J_{\xi, m}}(\mathbf{x}))^{2} \leq D_{1} \mathrm{E}(y(\mathbf{x})-y_{J^{*}_{m}}(\mathbf{x}))^{2},
\end{align}
and
\begin{align}
\label{ts19}
D_{2} m^{-2\gamma+1} \leq \mathrm{E}(y(\mathbf{x})-y_{J^{*}_{m}}(\mathbf{x}))^{2} \leq D_{3}m^{-2\gamma+1}.
\end{align}
\end{lam}

\noindent
{\bf Proof.}
By Lemmas \ref{l1} and \ref{l2}, it follows that
for all $1\leq m \leq (1-\epsilon)p$,
\begin{align*}
\begin{split}
& G_{1}m^{-2\gamma+1} \geq
\mathrm{E}(y(\mathbf{x})-y_{J_{\xi, m}}(\mathbf{x}))^{2} \geq \mathrm{E}(y(\mathbf{x})-y_{J^{*}_{m}}(\mathbf{x}))^{2} \\
& \geq \lambda_{1} \sum_{j \notin J^{*}_{m}} \beta^{*^2}_{j}  \geq \lambda_{1} \sum_{j \notin J^{o}_{m}} \beta^{*^2}_{j} \geq
\lambda_{1} L^{2} \sum_{j=m+1}^{p}j^{-2\gamma} \geq \lambda_{1} L^{2} \underline{d} m^{-2\gamma+1},
\end{split}
\end{align*}
where $J^{o}_{m}$ is the index set corresponding to $\{\beta^{2}_{(1)}, \cdots, \beta^{2}_{(m)}\}$
and $\underline{d}>0$ depends only on $\gamma$ and $\epsilon$.
These inequalities lead immediately to \eqref{ts18} and \eqref{ts19}.
\hfill$\Box$

\vspace{0.1cm}
\noindent
{\bf Remark A.1.}
Theorem 2.1 of Temlyakov (1998)
shows that a near best $m$-term approximation
can be realized by a greedy-type algorithm
under a basis $L_p$-equivalent to the Haar basis.
Since the Haar basis yields an identity correlation matrix,
our correlation assumption, \eqref{ts10},
appears to be substantially weaker.
The performance of the $m$-term approximation of OGA has been investigated by Tropp (2004)
under a noise-free underdetermined system
and a condition on
the cumulative coherence function,
which requires that the atoms in the dictionary are `nearly' uncorrelated.
His approximation error for OGA
is larger than that of the best $m$-term approximation
by a factor of $(1+6m)^{1/2}$.
Suppose that \eqref{intro3} holds.
Then,
\begin{eqnarray*}
\label{rev0}
\lambda_{1} \sum_{j \notin J^{o}_{m}} \beta^{*^2}_{j}
\leq
\mathrm{E}(y(\mathbf{x})-y_{J^{*}_{m}}(\mathbf{x}))^{2}
\leq
\mathrm{E}(\sum_{j \notin J^{o}_{m}} \beta_{j}^{*}z_j)^2,
\end{eqnarray*}
which, together with \eqref{intro3} and Minkowski's inequality, yields
\begin{align}
\label{rev1}
\begin{split}
C_{1,\beta}\lambda_1L_1^2{\rm exp}(-2\beta m)
\leq
\mathrm{E}(y(\mathbf{x})-y_{J^{*}_{m}}(\mathbf{x}))^{2}
\leq
C_{2,\beta}U_1^2{\rm exp}(-2\beta m),
\end{split}
\end{align}
where $C_{1,\beta}\leq C_{2,\beta} $
are some positive constants depending on $\beta$.
On the other hand, the argument used to prove
\eqref{ts12} leads to
\begin{align}
\label{rev2}
\begin{split}
\mathrm{E}(y(\mathbf{x})-y_{J_{\xi, m}}(\mathbf{x}))^{2}
=O({\rm exp}(-mf_{{\rm oga}})),
\end{split}
\end{align}
where $f_{{\rm oga}}=\xi^2\lambda_1(L1/U_1)^2(1-{\rm exp}(-\beta))^2< 2\beta$.
Equations
\eqref{rev1} and \eqref{rev2}
suggest that the population OGA
and the best $m$-term approximation
in general do not share the same convergence rate in the exponential decay case.
To be as efficient as the
best $m$-term approximation, the
population OGA needs to run for
another $m(2\beta/f_{{\rm oga}}-1)$ iterations,
which is still of order $m$.

\section{Proof of \eqref{31}}
\label{app1}
Recall that
\eqref{ts8} implies
\eqref{ubc} with $C=M+1$.
This, \eqref{ts3}, and \eqref{ts10}
yield
\begin{align*}
\begin{split}
&\big|n^{-1}\sum_{t=1}^{n}\varepsilon^2_{t}(\hat{J}_m)-\mathrm{E}_n(\varepsilon^2(\hat{J}_m))\big|
=
\big|\sum_{\sharp(J)=m}
\big\{
n^{-1}\sum_{t=1}^{n} \varepsilon_{t}^{2}(J)-\mathrm{E}(\varepsilon^{2}(J))
\big\}I_{\{\hat{J}_m=J\}}\big|\\
&\leq
\sum_{\sharp(J)=m}
\big\{\sum_{i=1}^{p}\sum_{l=1}^{p}
|\beta_i^{*}-\beta_{i}^{*}(J)||\beta_l^{*}-\beta_{l}^{*}(J)|
|n^{-1}\sum_{t=1}^{n}z_{ti}z_{tl}-\rho_{il}|
\big\}I_{\{\hat{J}_m=J\}} \\
&\leq
(M+1)^2\max_{1\leq i, l \leq p}|n^{-1}\sum_{t=1}^{n}z_{ti}z_{tl}-\rho_{il}|
\sum_{\sharp(J)=m}\big(\sum_{j \notin J}|\beta_i^*|\big)^2I_{\{\hat{J}_m=J\}} \\
&\leq
C_{\gamma}^2(M+1)^2R_{1,p}
\sum_{\sharp(J)=m}\big(\sum_{j \notin J}\beta_i^{*^2}\big)^{(2\gamma-2)/(2\gamma-1)}I_{\{\hat{J}_m=J\}} \\
&\leq
C_{M, \gamma, \lambda_1}R_{1,p}
\sum_{\sharp(J)=m}\big\{\mathrm{E}(\varepsilon^{2}(J))\big\}^{(2\gamma-2)/(2\gamma-1)}I_{\{\hat{J}_m=J\}} \\
&= C_{M, \gamma, \lambda_1}R_{1,p}
\big\{\mathrm{E}_n(\varepsilon^{2}(\hat{J}_m))\big\}^{(2\gamma-2)/(2\gamma-1)}.
\end{split}
\end{align*}
Thus, \eqref{31} follows.

\vspace{0.5cm}

\centerline{\bf SUPPLEMENTARY MATERIAL}
\vspace{0.1cm}
{\bf Supplement to ``Model selection for high-dimensional linear regression with dependent observations''}.
The supplementary material contains the proofs
of \eqref{rev32}, \eqref{332},
\eqref{ts-1}, \eqref{ts28}, \eqref{3-1}, \eqref{32}--\eqref{331}, \eqref{36},
and Lemma \ref{l2},
and a simulation study to demonstrate the performance of
OGA+HDAIC under a high-dimensional ARX model whose
$\bm{\Gamma}$
obeys $\lambda_{\max}(\bm{\Gamma}) \to \infty$
and \eqref{ts10}.

\section*{Acknowledgments}
I thank Hai-Tang Chiou, Hsueh-Han Huang, and Tze Leung Lai for helpful suggestions on earlier versions of this paper.
I would also like to thank an Associate Editor and two
anonymous referees for insightful and constructive comments.


\begin{thebibliography}{9}
\bibitem[AW(2015)] {AW2015} Adamczak, R. and Wolff, P. (2015).
Concentration inequalities for non-Lipschitz functions with bounded derivatives of higher order.
{\it Probab. Theory Relat. Fields} {\bf 162}, 531--586.
%
\bibitem[Basu and Michailidis(2015)] {BM2015} Basu, S.  and Michailidis, G. (2015).
Regularized estimation in sparse high-dimensional
time series models. {\it Ann. Statist.} {\bf 43}, 1535--1567.
%
\bibitem[Baxter(1962)] {B1962} {Baxter, G} (1962).
An asymptotic result for the finite predictor.
{\it Math. Scand.} {\bf 5}, 261--266.
%
\bibitem[Berk(1974)] {B1974} {Berk, K. N.} (1974).
Consistent autoregressive spectral estimates. {\it Ann. Statist.} {\bf 2}, 489--502.
%
\item[] \textrm{Bickel, P., Ritov, Y.} and \textrm{Tsybakov, A.} (2009).
Simultaneous analysis of Lasso and Dantzig selector.
\textit{Ann. Statist.}
\textbf{37}, 1705-1732.
%
\item[] \textrm{B\"{u}hlmann, P.} (2006).
Boosting for high-dimensional linear models.
\textit{Ann. Statist.}
\textbf{34}, 559-583.
%
\item[] \textrm{Candes, E. J.} and \textrm{Tao, T.} (2007).
The Dantzig selector: statistical estimation when $p$ is much larger than $n$.
\textit{Ann. Statist.}
\textbf{35}, 2313-2351.
%
\item[] \textrm{Chen, J.} and \textrm{Chen, Z.} (2008).
Extended Bayesian information criteria for model selection with large model spaces.
\textit{Biometrika}
\textbf{95}, 759-771.
%
\item[] \textrm{Ding, J.}, \textrm{Tarokh, V.} and \textrm{Yang, Y.}
Bridging AIC and BIC: a new criterion for autoregression.
\textit{IEEE Trans. Inform. Theory}
\textbf{64}, 4024--4043.
%
\item[] \textrm{Fan, J.} and \textrm{Lv, J.} (2008).
Sure independence screening for ultra-high dimensional feature space (with discussion).
\textit{J. Roy. Statist. Soc. Ser. B}
\textbf{70}, 849-911.
%
\item[] \textrm{Gao, F. , Ing, C.-K.} and \textrm{Yang, Y.} (2013).
Metric entropy and sparse linear approximation of $l_q$-Hulls for $0 < q \leq 1$.
\textit{J. Approx. Theory}
\textbf{166}, 42-55.
%
%
\bibitem[hi(2019)]{hI2019}
{Huang, H.-H.} and {Ing, C.-K.} (2019).
Concentration inequalities for sample higher order moments of stationary linear processes.
{\it Technical Report}, Institute of Statistics, National Tsing Hua University.
%
\bibitem[Ing(2007)]{Ing2007} {Ing, C.-K.} (2007). Accumulated prediction errors, information criteria and optimal forecasting for autoregressive
time series. {\it Ann. Statist.} {\bf 35}, 1238--1277.
%
\bibitem[Ing(2018)]{Ing2019} {Ing, C.-K.} (2019).
Supplement to ``Model selection for high-dimensional linear regression with dependent observations.''
%
\item[] \textrm{Ing, C.-K.} and \textrm{Lai, T. L.} (2011).
A stepwise regression method and consistent model
selection for high-dimensional sparse linear models. \textit{
Statistica Sinica} \textbf{ 21}, 1473--1513.
%
\item[] \textrm{Ing, C.-K., Lai, T. L., Shen, M, Tsang, K. W.} and {Yu, S.-H.} (2017).
Multiple testing in regression models with applications to fault diagnosis in big data era.
\textit{Technometrics} {\bf 59}, 351-360.
%
%
\item[] \textrm{Negahban, S. N., Ravikumar, P., Wainwright, M. J.} and \textrm{Yu, B.} (2012). A unified
framework for high-dimensional analysis of $M$-estimators with decomposable
regularizers. \textit{Statist. Sci.} {\bf 27}, 538--557.
%
\bibitem[Pourahmadi(1989)]{P1989} {Pourahmadi, M.} (1989). On the convergence of finite linear predictors of stationary processes.
{\it J. Multivariate Anal.} {\bf 30}, 167--180.
%
\item[] \textrm{Rudelson, M.} and \textrm{Vershynin, R.} (2013). Hanson-Wright inequality and sub-Gaussian concentration.
{\it Electron. Commun. Probab.} {\bf 18}, no. 82, 9.
%
\bibitem[Shibata(1980)]{Shibata1980} {Shibata, R.} (1980). Asymptotically efficient selection of the order of the model
for estimating parameters of a linear process. {\it Ann. Statist.} {\bf 8}, 147--164.
%
\item[] \textrm{Temlyakov, V. N.} (1998). The best $m$-term approximation and greedy algorithms.
\textit{Adv. Comput. Math.} \textbf{8}, 249--265.
%
\item[] \textrm{Temlyakov, V. N.} (2000).
Weak greedy algorithms.
\textit{Adv. Comput. Math.} \textbf{12}, 213--227.
%
\item[] \textrm{Temlyakov, V. N.} (2015). Greedy approximation in convex optimization.
{\it Constr. Approx.} {\bf 41}, 269--296.
%
\item[] \textrm{Tibshirani, R.} (1996).
Regression shrinkage and selection via the Lasso.
\textit{J. Roy. Statist. Soc. Ser. B}
\textbf{58}, 267-288.
%
\item[] \textrm{Tropp, J. A.} (2004).
Greed is good: Algorithmic results for sparse approximation.
\textit{IEEE Trans. Inform. Theory}
\textbf{50}, 2231-2242.
%
\item[] \textrm{Wang, H.} (2009).
Forward regression for ultra-high dimensional variable screening
\textit{J. Amer. Statist. Assoc.}
\textbf{104}, 1512-1524.
%
\item[]\textrm{Wang, Z., Paterlini, S., Gao, F.} and \textrm{Yang, Y.} (2014)
Adaptive Minimax Regression Estimation over Sparse $l_q$-Hulls.
\textit{J. Mach. Learning Res.}
\textbf{15}, 1675--1711.
%
\bibitem[Wu and Wu(2016)]{Wu16} Wu, W. B. and Wu, Y. N. (2016).
Performance Bounds for Parameter Estimates of High-dimensional
Linear Models with Correlated Errors. {\it Electron. J. Statist.} \textbf{10}, 352--379.
%
\bibitem[Yang(2007)]{Yang2007} {Yang, Y.} (2007). Prediction/estimation with simple linear model: Is it really that simple?
{\it Economet. Theory} {\bf 23}, 1--36.
%
\item[] \textrm{Zhang, C.-H.} (2010).
Nearly unbiased variable selection under minimax concave penalty.
\textit{Ann. Statist.}
\textbf{38}, 894--942.
%
\bibitem[Zhang and Yang (2015)]{ZhangYang2015} {Zhang, Y.} and {Yang, Y.} (2015).
Cross-validation for selecting a model selection procedure. {\it
J. Econometrics} {\bf 187}, 95--112.
%
\item[] \textrm{Zhao, P.} and \textrm{Yu, B.} (2006).
On model selection consistency of Lasso.
\textit{J. Machine Learning Res.}
\textbf{7}, 2541-2563.









\end{thebibliography}
\end{document}